\documentclass[letterpaper]{article} % DO NOT CHANGE THIS
\usepackage{aaai2026}  % DO NOT CHANGE THIS
\usepackage{times}  % DO NOT CHANGE THIS
\usepackage{helvet}  % DO NOT CHANGE THIS
\usepackage{courier}  % DO NOT CHANGE THIS
\usepackage[hyphens]{url}  % DO NOT CHANGE THIS
\usepackage{graphicx} % DO NOT CHANGE THIS
\usepackage{natbib}  % DO NOT CHANGE THIS AND DO NOT ADD ANY OPTIONS TO IT
\usepackage{caption} % DO NOT CHANGE THIS AND DO NOT ADD ANY OPTIONS TO IT
\usepackage{algorithm}
\usepackage{algorithmic}
\usepackage{amsmath}
\usepackage{amssymb}
\usepackage{booktabs}
\usepackage{colortbl}
\usepackage[table]{xcolor}
\usepackage{subcaption}
\usepackage{bibunits}

\usepackage{newfloat}
\usepackage{listings}
\DeclareCaptionStyle{ruled}{labelfont=normalfont,labelsep=colon,strut=off} % DO NOT CHANGE THIS
\floatstyle{ruled}
\newfloat{listing}{tb}{lst}{}
\floatname{listing}{Listing}
\title{Ultralight Polarity-Split Neuromorphic SNN for Event-Stream Super-Resolution}

\author{
    Chuanzhi Xu\thanks{Corresponding author.}, 
    Haoxian Zhou, 
    Langyi Chen, 
    Yuk Ying Chung, 
    Qiang Qu
}
\affiliations{
    School of Computer Science, The University of Sydney, NSW, Australia \\
    chuanzhi.xu@sydney.edu.au
}

\makeatletter
\def\ps@accepted{%
  \def\@oddhead{%
    \hfil
    \Large\textcolor{gray}{This paper has been accepted to AAAI 2026}%
    \hfil
  }%
  \def\@evenhead{\@oddhead}%
  \def\@oddfoot{}%
  \def\@evenfoot{}%
}
\makeatother

\begin{document}

\maketitle
\thispagestyle{accepted}
\begin{abstract}
Event cameras offer unparalleled advantages such as high temporal resolution, low latency, and high dynamic range. However, their limited spatial resolution poses challenges for fine-grained perception tasks. In this work, we propose an ultra-lightweight, stream-based event-to-event super-resolution method based on Spiking Neural Networks (SNNs), designed for real-time deployment on resource-constrained devices. To further reduce model size, we introduce a novel Dual-Forward Polarity-Split Event Encoding strategy that decouples positive and negative events into separate forward paths through a shared SNN. Furthermore, we propose a Learnable Spatio-temporal Polarity-aware Loss (LearnSTPLoss) that adaptively balances temporal, spatial, and polarity consistency using learnable uncertainty-based weights. Experimental results demonstrate that our method achieves competitive super-resolution performance on multiple datasets while significantly reducing model size and inference time. The lightweight design enables embedding the module into event cameras or using it as an efficient front-end preprocessing for downstream vision tasks.
\end{abstract}


\section{Introduction}
Event cameras, also known as neuromorphic cameras or dynamic vision sensors (DVS), are asynchronous sensors that respond to changes in scene brightness. When the brightness change at a given pixel exceeds a certain threshold, an event is triggered and recorded as a tuple $(x_k, y_k, t_k, p_k)$, where $(x_k, y_k)$ denotes the spatial coordinates of the pixel, $t_k$ is the precise timestamp, and $p_k \in \{+1, -1\}$ represents the polarity of brightness change \cite{Intro3, Survey}. The continuous sequence of $N$ events forms an event stream, which can be represented as:
\begin{equation}
EventStream = \left\{(t_k, x_k, y_k, p_k)\right\}_{k=1}^N.
\end{equation}

% The continuous sequence of $N$ events forms an event stream, which can be represented as:
% \begin{equation}
% EventStream = \left\{(t_k, x_k, y_k, p_k)\right\}_{k=1}^N.
% \end{equation}

Due to this asynchronous and sparse sensing mechanism, event streams exhibit several unique advantages over traditional frame-based RGB data, including high temporal resolution, high dynamic range (HDR), low latency, and negligible motion blur~\cite{Intro3, Survey12}. These properties make them especially promising for applications in 3D reconstruction~\cite{Intro2}, high-speed robotics~\cite{Survey23}, AR/VR~\cite{Intro4}, and autonomous driving~\cite{Survey22}.

\begin{figure}[t]
    \centering
    \includegraphics[width=\linewidth]{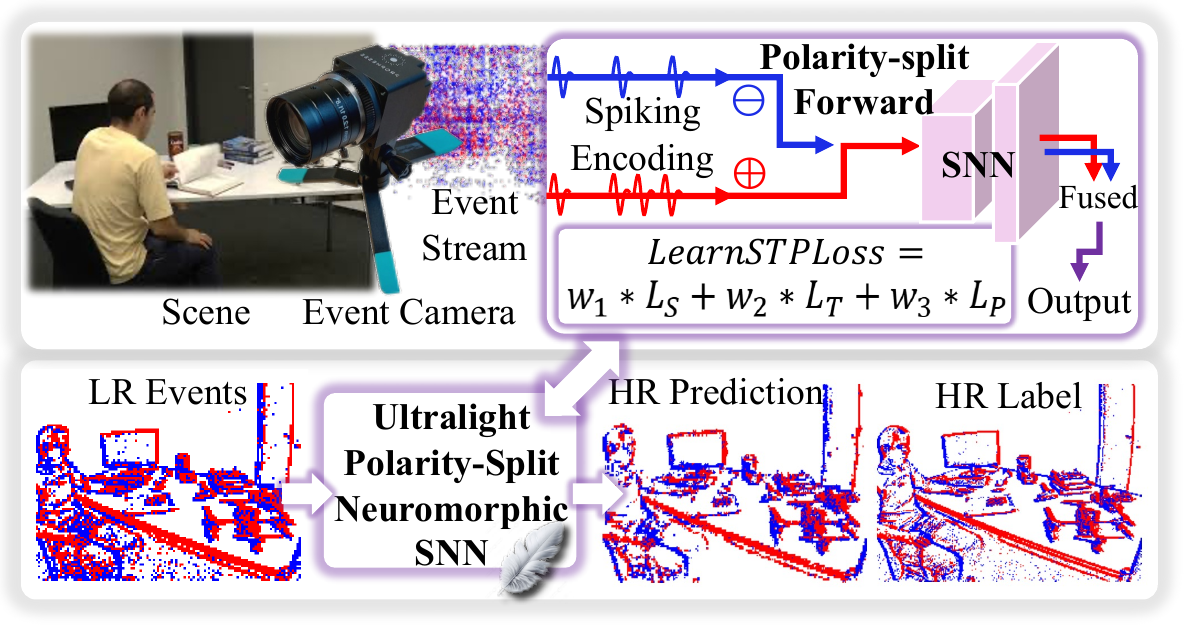}
    \caption{An overview of event stream super-resolution with Ultralight Polarity-Split Neuromorphic SNN.}
    \label{fig:intro}
\end{figure}

The spatial resolution of most commercially available event cameras ($\leq640\times480$) remains significantly lower than that of frame-based cameras \cite{SurveyEventCameraModel}. Although high-resolution event sensors such as the Sony IMX646 ($1280\times720$) have been developed \cite{SurveyEventCameraModel}, achieving higher resolution at the hardware level introduces increased power consumption and cost~\cite{EF4}. Moreover, recent studies suggest that the motivation for developing higher-resolution event cameras may be limited~\cite{Intro1}. In extreme conditions such as low-light scenes or high-speed motion, high-resolution event cameras may perform worse, since they elevate per-pixel event rates, which in turn amplify temporal noise~\cite{Intro1}. On the other hand, more studies show that under standard conditions, low-resolution event data limits fine-grained perception and downstream performance, while inputting high-resolution events can improve tasks like object recognition and image/video reconstruction~\cite{ES1, EF4, EF5, EF6}.

Event stream super-resolution (EventSR) emerges as the only technological direction that reconciles both perspectives: it aims to reconstruct high-resolution (HR) events from low-resolution (LR) input, as shown in Figure \ref{fig:intro}. Without the need to develop costly high-resolution event cameras, researchers can super-resolve event data in appropriate scenarios to achieve improved downstream task performance.

\begin{figure}[t]
    \centering
    \includegraphics[width=\linewidth]{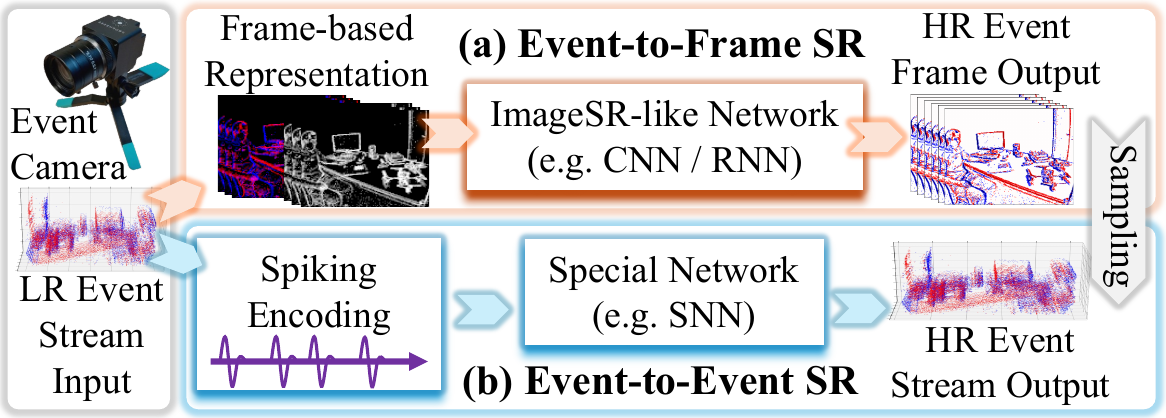}
\caption{Event-to-frame SR compresses the temporal dimension in exchange for spatial awareness and typically relies on heavier image SR models. Event-to-event SR requires specialized networks to directly generate asynchronous event streams, maintaining the nature of events.}
    \label{fig:compare}
\end{figure}

Existing event super-resolution methods can be broadly categorized into two types, referring to Figure \ref{fig:compare} and Section \ref{RelatedWork}. Event-to-frame super-resolution converts event streams into frame-based representations such as event stacks or event count maps~\cite{EF1, EF3, EF6}, and then applies reconstruction algorithms or networks similar to those used in image super-resolution. The frame-based output is then uniformly or randomly sampled on the temporal dimension to produce high-resolution event streams. As many downstream vision tasks still require frame-based inputs to interface with processing modules like CNNs, the loss of temporal precision caused during this process is often considered acceptable. The second type is event-to-event stream-based super-resolution, which directly reconstructs high-resolution event streams without any temporal sampling~\cite{ES1}. It preserves the asynchronous and temporally precise nature of event data, while recovering such temporally asynchronous events requires specialized networks such as Spiking Neural Networks (SNNs)~\cite{ES1}.

%Existing event stream super-resolution methods can be broadly categorized into two classes. The first is frame-based event super-resolution, which stacks event streams into synchronous representations such as event stacks or event count maps~\cite{EF1, EF3, EF6}, and then applies reconstruction algorithms similar to those used in image super-resolution. The frame-based output is then uniformly or randomly sampled on the temporal dimension to produce high-resolution event streams. As many downstream vision tasks still require frame-based inputs to interface with processing modules like CNNs, the loss of temporal precision introduced during this process is often considered acceptable. The second class is end-to-end event stream super-resolution, which directly reconstructs high-resolution event streams without any temporal sampling~\cite{ES1, ES2}. This approach preserves the asynchronous and temporally precise nature of event data, while recovering such temporally asynchronous events requires specialized architectures such as Spiking Neural Networks (SNNs)~\cite{ES1}.

However, for the motivation of embedding a controllable super-resolution module on lightweight event cameras, only the event-to-event super-resolution is suitable. An event super-resolution module should not assume the processing of subsequent vision tasks. It is necessary to recover high-temporal-resolution event stream data in this case.

More importantly, an ideal event super-resolution module should be \textbf{lightweight}, \textbf{energy-efficient}, and capable of \textbf{real-time} processing. It is impractical to allocate a high-end chip solely for running a heavy model on an event camera, and it is also undesirable to dedicate substantial computational resources to a super-resolution module within a vision processing pipeline as a step of preprocessing. However, recent methods have made this module still heavy~\cite{EF5, EF6}.

Therefore, we propose an ultra-lightweight, real-time, stream-based event-to-event super-resolution network based on SNN. Figure \ref{fig:intro} provides an overview of our method. It not only improves super-resolution accuracy but also significantly reduces model parameters, further accelerating inference. This makes it feasible to embed lightweight event super-resolution modules into event cameras or use them as energy-efficient visual preprocessing units. 

Our contributions can be summarized as follows:
\begin{itemize}
    \item We propose an ultra-lightweight, SNN-based event-to-event super-resolution network that achieves higher super-resolution accuracy while enabling real-time deployment on resource-constrained devices.
    \item We introduce a neuromorphic forward propagation strategy named Dual-Forward Polarity-Split Event Encoding, which further reduces the model size by half and improves the spatio-temporal precision.
    \item By integrating our proposed Learnable Spatio-temporal Polarity-aware Loss (LearnSTPLoss), our method achieves superior reconstruction accuracy across multiple datasets, and the super-resolved event streams further enhance downstream tasks of object recognition and image reconstruction.
\end{itemize}

\section{Related Work} \label{RelatedWork}

\subsection{Event-to-Frame Super-Resolution}
%Most event super-resolution methods rely on converting the event stream into an intermediate frame-based representation, such as the event count maps or event stacks~\cite{EF1, EF3, EF6}, which can be processed by traditional algorithms and neural networks similar to those used in image super-resolution. The reconstructed frames are then uniformly or randomly sampled on the temporal dimension to recover the HR event stream, which often results in reduced temporal resolution. As many downstream vision tasks still require frame-based inputs to interface with processing modules like CNNs, the loss of temporal precision introduced during this process is often considered acceptable. However, many of them are too heavy and suffer from low temporal precision to be suitable for on-device deployment within event cameras.

Before 2020, event-to-frame super-resolution was mainly based on mathematical modeling. Li et al. were the first to address this task \cite{EF1}. They modeled the event stream at each pixel as a non-homogeneous Poisson process, using event count maps to recover the number of events in the spatial domain and spatiotemporal filters to estimate the event rate function in the temporal domain. Later, Wang et al. proposed the Guided Event Filtering method \cite{EF2}, which jointly filtered frame images and event data to improve spatial denoising in event super-resolution.

Subsequently, deep learning-based methods were developed. In 2021, Duan et al. introduced EventZoom \cite{EF3}, a method based on a 3D U-Net architecture with an event-to-image module to leverage high-resolution image features, achieving better results than previous non-learning approaches. In 2022, Weng et al. proposed RecEvSR \cite{EF4}, which used an RNN with a spatial attention mechanism to enhance detail in key areas, showing strong performance for high upsampling rates.

More recent works have explored separating polarities and feeding them into different branches of a network, with interaction modules to exchange information. In 2024, Huang et al. proposed BMCNet \cite{EF5}, a bilateral mining and complementary fusion framework that extracts positive and negative polarity features separately, then fuses them to improve detail recovery. Liang et al. proposed RMFNet \cite{EF6}, which combines a dual-branch architecture with a recursive structure to enhance temporal modeling across frames and further improve performance.

\begin{figure*}[ht]
    \centering
    \includegraphics[width=\linewidth]{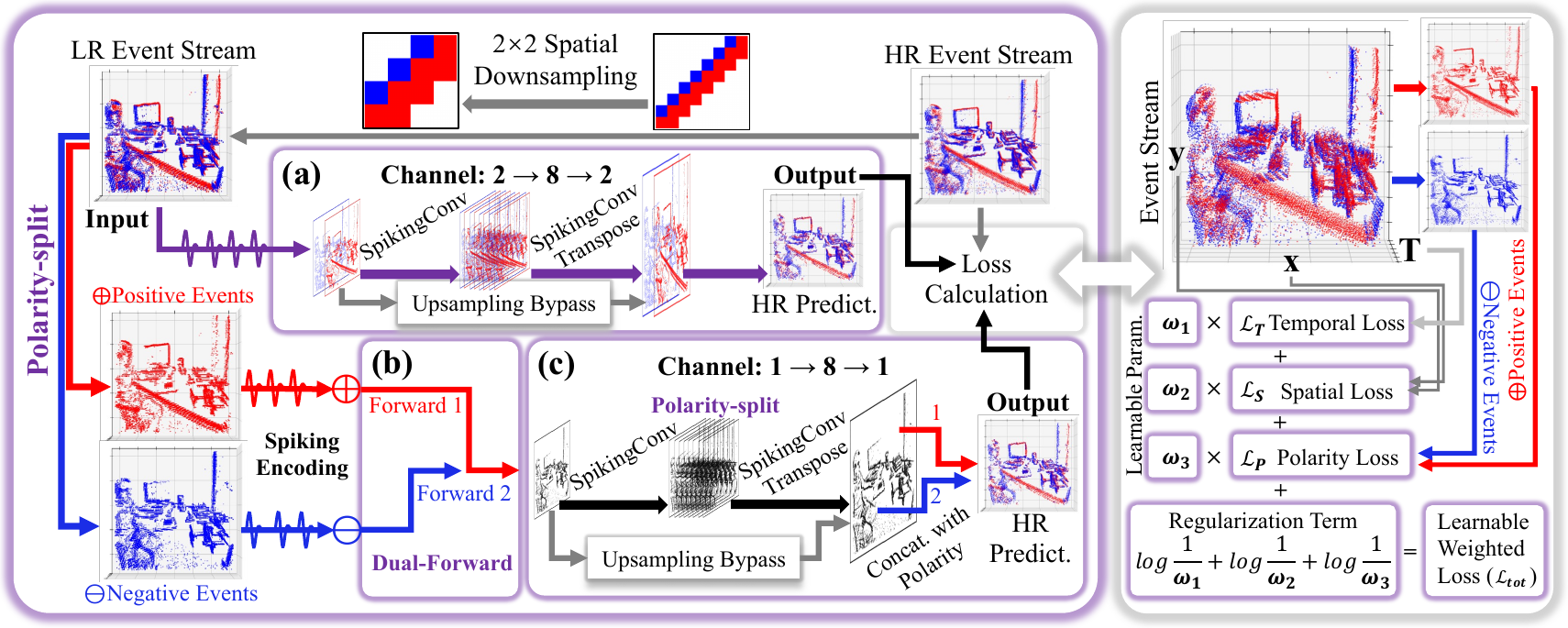}
    \caption{Architecture of key modules in our proposed event super-resolution method. Left: Main architecture of our method. Right: Composition of the Learnable Spatio-temporal Polarity-aware Loss function. (a) Dual-layer EventSR Network. (b) Dual-Forward Polarity-Split Event Encoding Strategy. (c) Ultra-lightweight Polarity-Split EventSR Network.}
    \label{fig:method}
\end{figure*}

\begin{figure}[h]
    \centering
    \includegraphics[width=\linewidth]{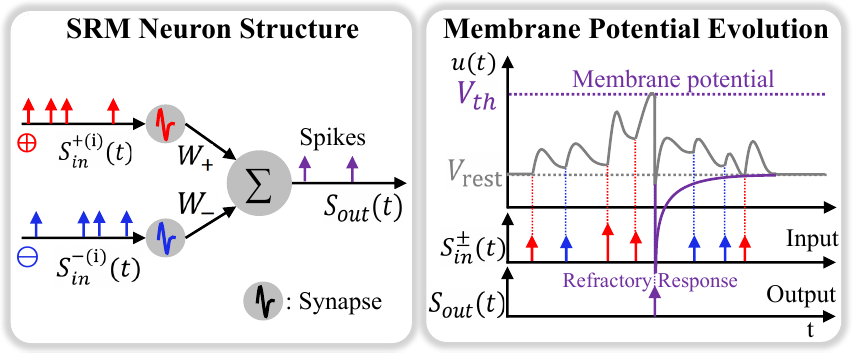}
    \caption{SRM-based spiking neuron structure and spike triggering mechanism.}
    \label{fig:SNN}
\end{figure}

\subsection{Event-to-Event Super-Resolution}
Only a few methods are event-to-event super-resolution that preserve the asynchronous and temporally precise nature of the original event stream.

In 2021, Li et al. introduced an innovative approach using SNNs for the event super-resolution task \cite{ES1}. SNNs can directly process raw event streams and output high-resolution asynchronous event streams, maintaining the original temporal characteristics. In addition, they designed spatiotemporal constraints to enable the network to learn both spatial and temporal distributions of events. This method serves as the primary baseline for our work.

\begin{figure*}[th]
    \centering
    \includegraphics[width=0.7\linewidth]{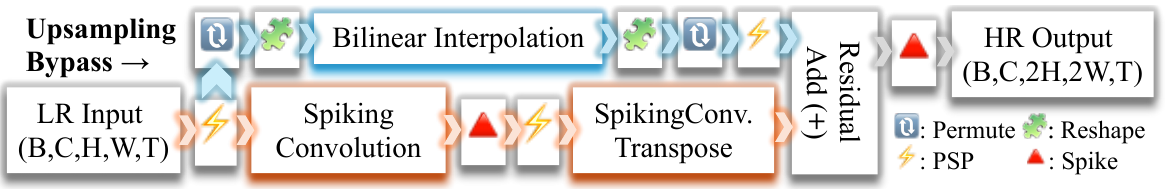}
    \caption{Illustration of the main network path and the PSP-based residual upsampling bypass.}
    \label{fig:model}
\end{figure*}

\section{Methods}

\subsection{SRM-based Spiking Neuron Encoding}
We adopt a Spiking Neural Network (SNN) based on the Spike Response Model (SRM) \cite{SRM} to address the task of event stream super-resolution. Unlike conventional artificial neurons that rely on continuous and differentiable activation functions, SRM-based spiking neurons encode and transmit information through temporal spike trains, simulating biological neural behavior, as shown in Figure \ref{fig:SNN}.

Let $s_\text{in}^{(i)}(t)$ and $s_\text{out}(t)$ denote the input and output neuron spike trains, respectively. Each neuron maintains an internal membrane potential $u(t)$, which integrates incoming spikes via synaptic weights. The contribution from each input spike train is first processed through a spike response kernel $\epsilon(t)$ to generate a Post Synaptic Potential (PSP), representing the temporal influence of a spike. This PSP is then scaled by a synaptic weight $w^{(i)}$ before being accumulated into the membrane potential. The PSP is computed as a temporal convolution of $s_\text{in}^{(i)}(t)$ with $\epsilon(t)$. When $u(t)$ surpasses a threshold $V_\text{th}$, the neuron emits an output spike and triggers a refractory mechanism that suppresses but does not reset the membrane potential. This mechanism is controlled by a time constant $\tau_\text{ref}$ and suppression coefficient $\lambda$, which together determine the duration and strength of the inhibition. The membrane potential dynamics are given by:
\begin{equation}
u(t) = \sum_{i} w^{(i)} \cdot (\epsilon(t) * s_\text{in}^{(i)}(t)) + (\gamma(t) * s_\text{out}(t)),
\end{equation}
where $*$ denotes convolution, and $\gamma(t)$ is the refractory response kernel.

Considering a feed-forward SNN consists of $L$ layers. The membrane potential and spiking output at layer $l+1$ are:
\begin{align}
u^{(l+1)}(t) &= \mathbf{W}^{(l)} \cdot \left[ \epsilon(t) * s^{(l)}(t) \right] + \gamma(t) * s^{(l+1)}(t), \\
s^{(l+1)}(t) &= \sum_{t_k \in \{t \mid u^{(l+1)}(t) = V_\text{th} \}} \delta(t - t_k),
\end{align}
where $\mathbf{W}^{(l)}$ denotes the synaptic weight matrix, and $\delta$ is the Dirac function representing spike times.

The specific forms of the response kernels are:
\begin{align}
\epsilon(t) &= \frac{t}{\tau_s} \exp\left(1 - \frac{t}{\tau_s}\right) \cdot \Theta(t), \\
\gamma(t) &= -\lambda \cdot \exp\left(-\frac{t}{\tau_r}\right) \cdot \Theta(t),
\end{align}
where $\tau_s$ and $\tau_r$ are time constants for the spike and refractory kernels, respectively, $\lambda$ is the configurable suppression coefficient, and $\Theta(t)$ is the Heaviside step function.

\subsection{Dual-layer EventSR SNN}
Most previous works tend to increase the complexity of neural architectures for event stream super-resolution to achieve better performance~\cite{EF5, EF6}. However, the baseline work has shown that for SNN-based event stream super-resolution, deeper networks may actually degrade performance~\cite{ES1}, which supports our motivation to further simplify the network design.

\begin{table}[h]
\centering
\resizebox{\linewidth}{!}{
\begin{tabular}{lcccc}
\toprule
\textbf{Layer} & \textbf{Channels} & \textbf{Kernel Size} & \textbf{Stride} & \textbf{Padding} \\
\midrule
SpikingConv.      & 2 $\Rightarrow$ 8 & $5\times5$ & 1 & 2 \\
SpikingConv. Transpose & 8 $\Rightarrow$ 2 & $2\times2$ & 2 & 0 \\
\bottomrule
\end{tabular}
}
\caption{Configuration of Dual-layer EventSR Network.}
\label{tab:SNN1}
\end{table}

We propose a compact yet effective SNN-based event stream super-resolution network, simply named \textbf{Dual-layer SNN} in this paper. As shown in Figure \ref{fig:method} and Table~\ref{tab:SNN1}, this network consists of two primary spiking convolution layers. The input and output of the network contain two channels, representing positive and negative event streams. The first layer extracts local spatiotemporal spike patterns and generates intermediate spike responses. The second layer upsamples the output of the first layer back to the original resolution. As shown in Figure \ref{fig:model}, it also incorporates an upsampling bilinear-interpolated PSP bypass to preserve fine details from the low-resolution input for final reconstruction.

\begin{table}[h]
\centering
\resizebox{0.85\linewidth}{!}{
\begin{tabular}{lcccccc}
\toprule
\textbf{Layer} & \boldmath$V_\text{th}$ & \boldmath$\tau_s$ & \boldmath$\tau_r$ & \boldmath$\lambda$ & \boldmath$\tau_\rho$ & \boldmath$\rho$ \\
\midrule
SpikingConv. & 30  & 1 & 1 & 1 & 1  & 10 \\
SpikingConv. Transpose & 100 & 4 & 4 & 1 & 10 & 100 \\
\bottomrule
\end{tabular}
}
\caption{Neuron parameter configuration. $\tau_\rho$ and $\rho$ control the time constant and scaling factor of the surrogate gradient used during backpropagation.}
\label{tab:neuron}
\end{table}

For neuron parameters, we adopt a configuration (see Table~\ref{tab:neuron}) that enables rapid response to sparse inputs in early layers and stronger integration in later layers. This helps the network effectively learn the mapping for temporal density restoration and spatial upsampling, without relying on deep layer stacking.

\subsection{Dual-Forward Polarity-split Event Encoding} \label{sec:strategy}
Recent studies have proposed using dual-branch networks to separately process positive and negative events \cite{EF5, EF6}, demonstrating that handling polarities separately leads to improved performance on event super-resolution. However, the additional network branches significantly increase the model's weight.

We novelly propose a polarity-aware event data forward strategy - Dual-Forward Polarity-Split Event Encoding (See Figure \ref{fig:method}b and Alg. \ref{stretegy}), simply named \textbf{Dual-Forward strategy} in this paper. Instead of treating the polarity channels as a single input, we decouple the input event stream tensors \( e^{\text{(in)}} \in \mathbb{R}^{2 \times H \times W \times T} \) into positive and negative streams:
\begin{equation}
e^{+} = e^{\text{(in)}}_{0,:,:,:}, \quad e^{-} = e^{\text{(in)}}_{1,:,:,:},
\end{equation}
where \( e^{+} \) and \( e^{-} \) denote the event tensors corresponding to positive and negative polarities, respectively.

These are then forwarded independently through the shared SNN-based network \( \mathcal{F}(\cdot; \theta) \):
\begin{equation}
\hat{e}^{+} = \mathcal{F}(e^{+}; \theta), \quad \hat{e}^{-} = \mathcal{F}(e^{-}; \theta).
\end{equation}

Then, integrating the two outputs as the final output:
\begin{equation}
\hat{e}^{\text{(out)}} = \text{Concat}(\hat{e}^{+}, \hat{e}^{-}) \in \mathbb{R}^{2 \times H' \times W' \times T}.
\end{equation}
This integrated output is used to calculate the loss, followed by a joint backpropagation.

This strategy offers many advantages. It removes the need for double-channel event handling within each SNN layer, significantly reducing model size and parameter count. In the experimental phase, running these two forward propagations concurrently also speeds up inference.
Moreover, it preserves polarity-specific spatiotemporal dynamics via dedicated forward paths while maintaining unified training through shared losses.

\begin{algorithm}[t]
\caption{Dual-Forward Polarity-Split Event Encoding}
\begin{algorithmic}[1]
\REQUIRE Input LR event stream \( e^{\text{(in)}} \)%, ground truth \( e^{\text{(out)}} \)
\STATE Split into polarity channels: \( e^{+} \gets e^{\text{(in)}}_{0,:,:,:} \), \( e^{-} \gets e^{\text{(in)}}_{1,:,:,:} \)
\STATE Forward pass separately/concurrently:
\STATE \quad $1st:$ \( \hat{e}^{+} \gets \mathcal{F}(e^{+}; \theta) \)
\STATE \quad $2nd:$ \( \hat{e}^{-} \gets \mathcal{F}(e^{-}; \theta) \)
\STATE Concatenate: \( \hat{e}^{\text{(out)}} \gets \text{Concat}(\hat{e}^{+}, \hat{e}^{-}) \)
\STATE Compute total loss: \( \mathcal{L}_\text{total} \gets \text{LearnSTPLoss}(e_\text{out}, e_\text{gt}) \)
\STATE Backpropagate: \( \theta \gets \theta - \nabla_\theta \mathcal{L}_\text{total} \)
\end{algorithmic}\label{stretegy}
\end{algorithm}

\begin{figure*}[t]
    \centering
    \includegraphics[width=\linewidth]{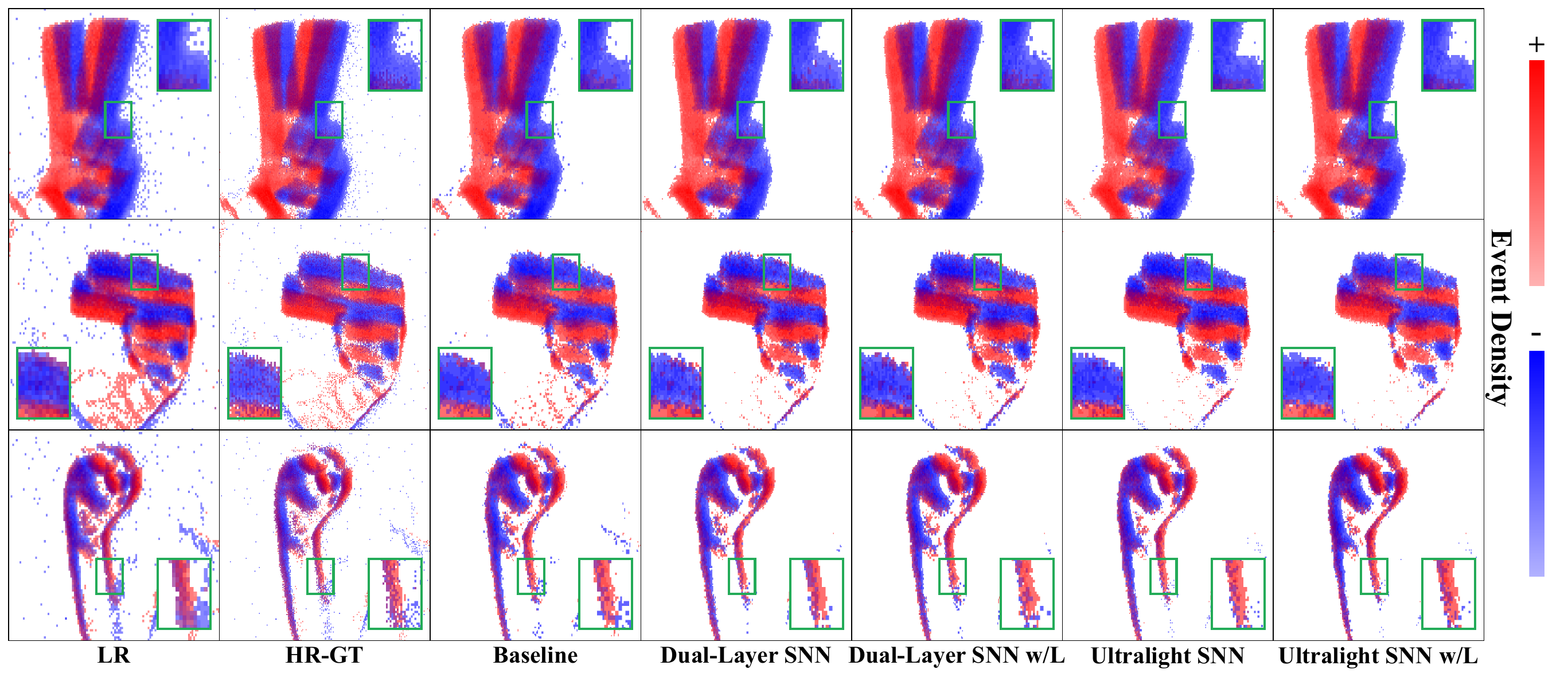}
    \caption{Visualizations on the ASL-DVS. Positive (red) and negative (blue) events are accumulated on each pixel.}
    \label{fig:ASL}
\end{figure*}

\begin{figure}[th]
    \centering
    \includegraphics[width=\linewidth]{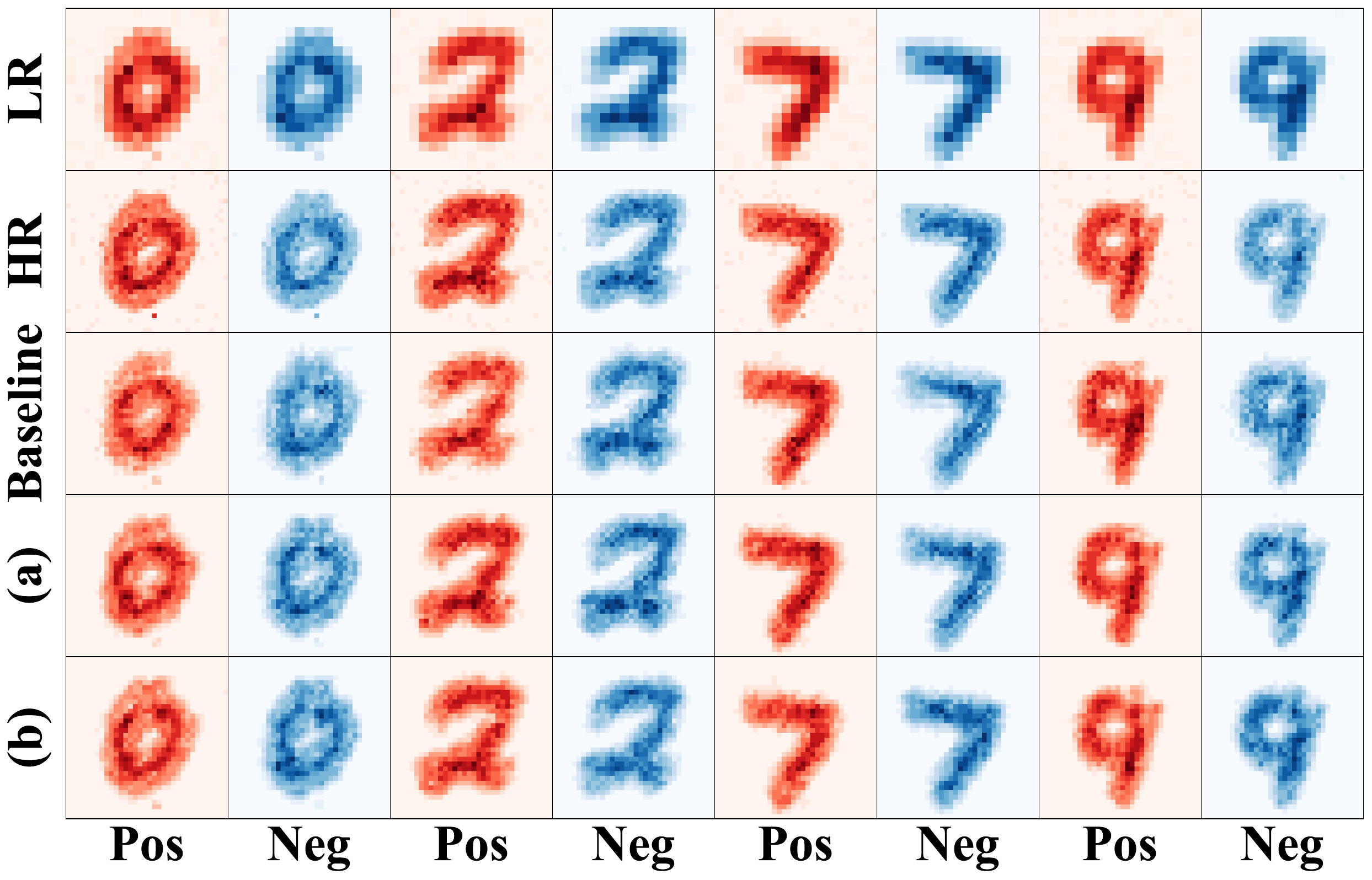}
    \caption{Visualizations on the N-MNIST, split by polarity. (a) Dual-Layer SNN w/Loss. (b) Ultralight SNN w/Loss.}
    \label{fig:NMIST}
\end{figure}

\subsection{Ultra-lightweight EventSR SNN}

We integrate the Dual-Forward Polarity-Split Event Encoding strategy into our dual-layer SNN event stream super-resolution network, simply named \textbf{Ultralight SNN} in this paper. As shown in Figure \ref{fig:method}c, since events are forward-propagated separately by polarity, the model only requires a single input and output channel (reduced from 2 to 1), with polarity information encoded within independent propagation groups. The model becomes significantly more lightweight. Moreover, this design helps spatio-temporal perception while introducing only a minor impact on polarity accuracy, as shown in the experimental results.

%\begin{table}[ht]
%\centering
%\caption{Configuration of Lightweight Polarity-split EventSR Network}
%\resizebox{\linewidth}{!}{
%\begin{tabular}{l|c|c|c|c}
%\hline
%\textbf{Layer} & \textbf{Channels} & \textbf{Kernel Size} & \textbf{Stride} & \textbf{Padding} \\
%\hline
%Spiking Conv      & 1 $\Rightarrow$ 8 & 5$\times$5 & 1 & 2 \\
%Spiking TransConv & 8 $\Rightarrow$ 1 & 2$\times$2 & 2 & -- \\
%\hline
%\end{tabular}
%}
%\end{table}
\begin{figure*}[htbp]
    \centering
    \includegraphics[width=\linewidth]{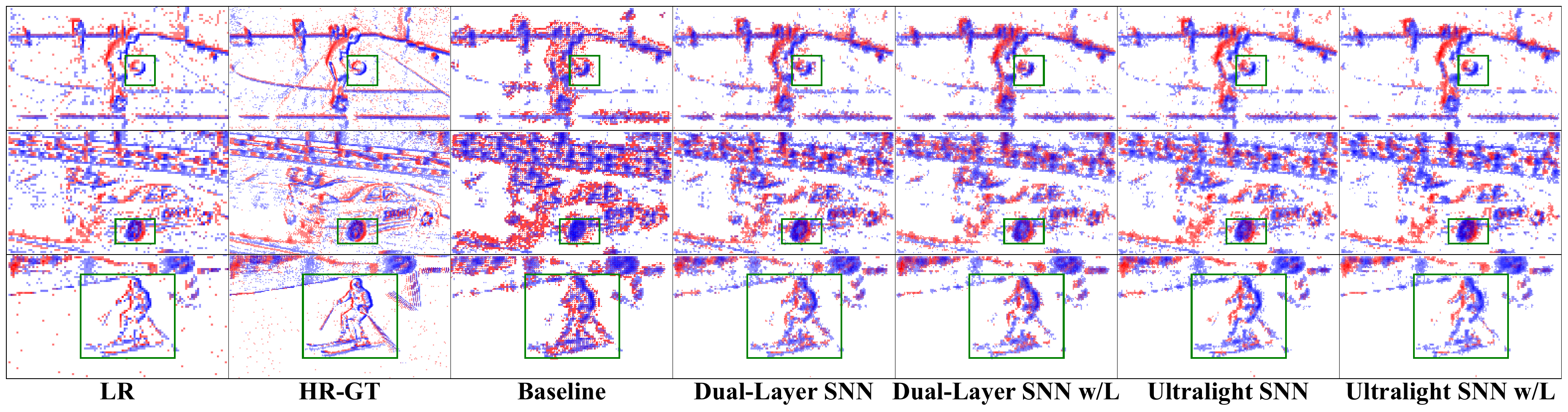}
    \caption{Visualizations on the EventNFS-real. (Zoom up to see the performance on edges.)}
    \label{fig:NFS}
\end{figure*}

\subsection{Learnable Spatio-temporal Polarity-aware Loss}

Inspired by previous work that performs the spatio-temporal learning~\cite{ES1} and weighted uncertainty loss \cite{LearnableLoss}, we extend the spatio-temporal only loss with polarity awareness and learnable adaptive weighting, namely \textbf{Learn}able \textbf{S}patio-\textbf{t}emporal \textbf{P}olarity-aware Loss (\textbf{LearnSTPLoss}), simply marked as "w/L" in this paper. Given the predicted spike output $e_\text{out}$ and the ground truth spike stream $e_\text{gt}$, we have:

\subsubsection{Temporal Loss.} $\mathcal{L}_\text{T}$ maintains temporal accuracy by dividing events into $T$ time windows and comparing them frame by frame.
\begin{equation}
\mathcal{L}_\text{T} = \frac{1}{T} \sum_{t=1}^T \left\| e_\text{out}(:, :, :, :, t) - e_\text{gt}(:, :, :, :, t) \right\|_2^2.
\end{equation}

\subsubsection{Spatial Loss.} $\mathcal{L}_\text{S}$ accumulates spike values over $B$ temporal bins $\mathcal{T}_b$ to measure deviation in spatial projections.
\begin{equation}
\mathcal{L}_\text{S} = \sum_{b=1}^{B} \left\| \sum_{t \in \mathcal{T}_b} e_\text{out}(:, :, :, :, t) - \sum_{t \in \mathcal{T}_b} e_\text{gt}(:, :, :, :, t) \right\|_2^2
\end{equation}

\subsubsection{Polarity-aware Loss.} To enhance polarity fidelity, we introduce a polarity-aware loss that evaluates the discrepancy across positive and negative event channels independently:
\begin{equation}
\mathcal{L}_\text{P} = \left\| e_\text{out}^{(+)} - e_\text{gt}^{(+)} \right\|_2^2 + \left\| e_\text{out}^{(-)} - e_\text{gt}^{(-)} \right\|_2^2.
\end{equation}

\subsubsection{Learnable Weighted Total Loss.} Preserving time, space, and polarity information is essential to the event stream super-resolution task. When using a combined loss function, previous methods often rely on manually set hyperparameters to weight each component \cite{ES1}. However, such manual tuning lacks adaptability across different datasets or model architectures. To address this issue, we introduce a learnable weighting mechanism based on the logarithm of variance, enabling the weights of each loss term to be jointly balanced and optimized with the network parameters during training. This learnable loss formulation is expected to improve super-resolution accuracy without compromising the model's lightweight design.

The final total loss can be calculated as:
\begin{equation}
\begin{split}
\mathcal{L}_\text{total} =\ & w_1 \cdot \mathcal{L}_\text{T} + w_2 \cdot \mathcal{L}_\text{S} + w_3 \cdot \mathcal{L}_\text{P} \\
&+ \log \frac{1}{w_1} + \log \frac{1}{w_2} + \log \frac{1}{w_3},
\end{split}
\end{equation}
where each weight $w_i$ represents a learnable parameter with uncertainty proxy $\sigma_i^2$:
\begin{equation}
    w_i = \exp(-\log \sigma_i^2).
\end{equation}
This ensures each weight remains positive, and the regularization term avoids collapsing weights to zero.

%During training, a single loss function is computed over the combined output:
%\begin{equation}
%\mathcal{L}_\text{total} = \mathcal{L}_\text{recon}(\hat{e}^{\text{(out)}}, e^{\text{(out)}}) + \lambda_\text{ecm} \cdot \mathcal{L}_\text{ecm}(\hat{e}^{\text{(out)}}, e^{\text{(out)}})
%\end{equation}
%where \( \mathcal{L}_\text{recon} \) is a reconstruction loss (e.g., MSE), and \( \mathcal{L}_\text{ecm} \) is a temporal block-based consistency loss.

\section{Experiments}

\subsection{Experimental Settings}

\subsubsection{Datasets.}
We evaluate our method on five datasets, as summarized in Table~\ref{Dataset}. The HR event streams correspond to the original event data in these datasets. To generate the LR event streams, we perform spatial downsampling by merging events within each 2×2 region using a stride of 2, which is shown in Figure \ref{fig:method}.

\subsubsection{Implementation Details.}
We implement our SNN architecture using SlayerSNN~\cite{SlayerSNN}, with a discretized simulation step size of 1 millisecond to update membrane potentials and determine spike firing. We train our model for 30 epochs using the Adam optimizer with an initial learning rate of 0.1. Table \ref{Dataset} provides the batch size for training each of the datasets. All experiments are conducted on an NVIDIA RTX 5090 GPU.

We conduct 2× spatial super-resolution experiments (i.e., doubling both Height (H) and Width (W)), which is sufficient to cover practical use cases for embedding super-resolution modules in event cameras.

\subsection{Evaluation Metrics}
To quantitatively evaluate the performance of event stream super-resolution, referring to our baseline \cite{ES1}, we implement a Root Mean Squared Error (RMSE) metric that considers both temporal (MSE$_T$) and spatial (MSE$_S$) precision. The predicted and ground-truth event streams are represented as four-column arrays $(t, x, y, p)$ and discretized into 4D voxel tensors of shape $(C{=}2, H, W, T)$, where the two channels correspond to positive and negative polarities. In the formulations below, $(i,j)$ index pixel locations, $t$ denotes the temporal coordinate within the interval $[T_0, T_1]$, $N_p$ is the number of active pixels, $k$ indexes the temporal bins, and $N_b$ is the total number of bins.

\begin{table}[t]
\centering
\resizebox{\linewidth}{!}{
\begin{tabular}{lccc}
\toprule
\textbf{Dataset} & \textbf{Train/Test} & \textbf{Sensor} & \textbf{bs} \\
\midrule
N-MNIST \cite{NMnist} & 60k / 10k & ATIS & 64 \\
CIFAR10-DVS \cite{Cifar} & 8.5k / 1.5k & DVS128 & 8 \\
ASL-DVS \cite{ASL} & 80.6k / 20.2k & DAVIS240C & 32 \\
EventCameraD. \cite{EventCameraDataset} & 10.1k / 1.6k & DAVIS240C & 64 \\
EventNFS-real \cite{EF3} & 74.2k / 10.5k & DAVIS346 & 64 \\
\bottomrule
\end{tabular}
}
\caption{Datasets and Batch Size (bs) information.}
\label{Dataset}
\end{table}

\begin{figure*}[ht]
    \centering
    \includegraphics[width=\linewidth]{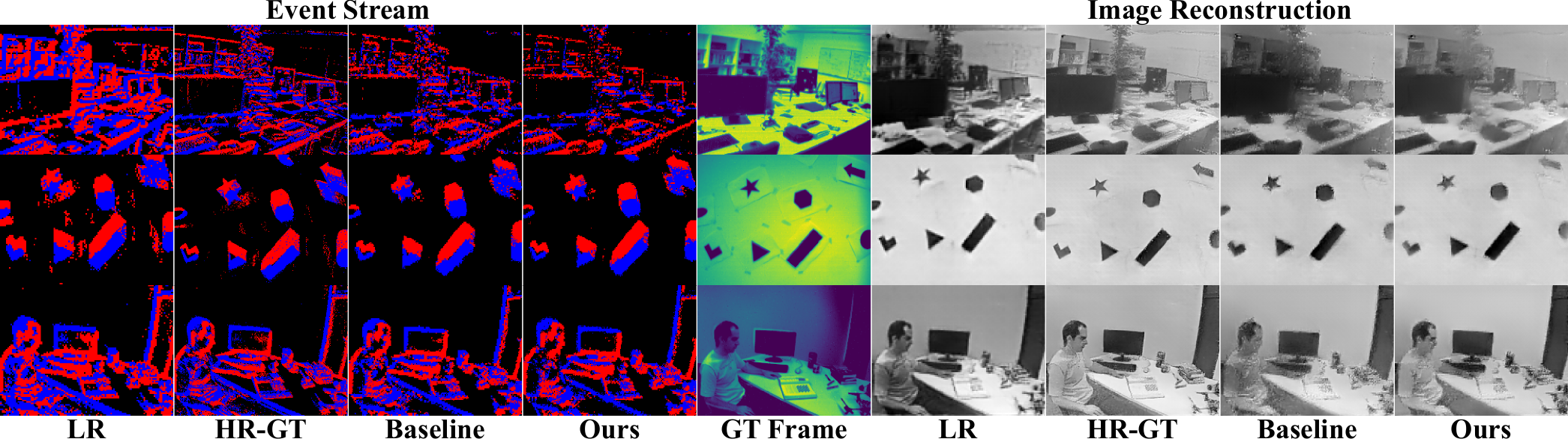}
    \caption{Visualizations of Image Reconstruction on Event Camera Dataset. Our method shown here is Ultralight SNN w/LearnSTPLoss, and it performs better on edges. (Zoom up to see the performance on edges.)}
    \label{fig:image_recon}
\end{figure*}

\begin{table*}[ht]
\small
\setlength{\tabcolsep}{2pt}
\centering
\resizebox{\textwidth}{!}{
\begin{tabular}{l|cccc|cccc|cccc|cccc|cccc}
\toprule
\textbf{Method} 
& \multicolumn{4}{c|}{\textbf{N-MNIST}} 
& \multicolumn{4}{c|}{\textbf{CIFAR10-DVS}} 
& \multicolumn{4}{c|}{\textbf{ASL-DVS}} 
& \multicolumn{4}{c|}{\textbf{Event Camera Dataset}} 
& \multicolumn{4}{c}{\textbf{EventNFS-real}} \\
 & RMSE$_\text{ST}$ & MSE$_\text{S}$ & MSE$_\text{T}$ & PA (\%) 
 & RMSE$_\text{ST}$ & MSE$_\text{S}$ & MSE$_\text{T}$ & PA (\%) 
 & RMSE$_\text{ST}$ & MSE$_\text{S}$ & MSE$_\text{T}$ & PA (\%) 
 & RMSE$_\text{ST}$ & MSE$_\text{S}$ & MSE$_\text{T}$ & PA (\%) 
 & RMSE$_\text{ST}$ & MSE$_\text{S}$ & MSE$_\text{T}$ & PA (\%) \\
\midrule
Baseline                    
& .2720 & .1440 & .2301 & 96.36 
& .1792 & .1011 & .1479 & 69.42 
& .2290 & .1055 & .2008 & 98.80 
& .3250 & .2753 & .1723 & 83.82
& .3448 & .2414 & .2460 & 76.69 \\
D-L
& .2621 & .1292 & .2280 & \underline{97.34} 
& .1781 & .1002 & .1473 & \underline{69.92} 
& .2250 & \underline{.1048} & .1975 & \underline{99.39} 
& .3165  & \underline{.2593} & .1697 & 89.06
& .3180  & .2160  & .2329 & 87.46 \\
D-L w/L            
& \textbf{.2607} & \textbf{.1277} & \textbf{.2272} & \textbf{97.87} 
& .1772 & .1017 & .1451 & \textbf{70.17} 
& .2250 & \textbf{.1038} & .1982 & \textbf{99.42} 
& .3130  & .2626 & \underline{.1675} & \underline{89.82}
& \underline{.3105} & .2148 & \underline{.2237} & 87.65 \\
U                   
& .2614 & .1289 & .2274 & 95.63 
& \underline{.1754} & \textbf{.0992} & \underline{.1446} & 68.28 
& \underline{.2247} & .1069 & \underline{.1961} & 98.87 
& \underline{.3128}  & .2626 & \textbf{.1674} & \textbf{89.85}
& .3113 & \underline{.2123} & .2270  & \underline{89.35} \\
U w/L          
& \underline{.2611} & \underline{.1283} & \underline{.2274} & 95.74 
& \textbf{.1747} & \underline{.0994} & \textbf{.1437} & 68.78 
& \textbf{.2236} & .1067 & \textbf{.1948} & 98.93 
& \textbf{.3117}  & \textbf{.2592} & .1699 & 89.03
& \textbf{.3061} & \textbf{.2118} & \textbf{.2202} & \textbf{89.96} \\
\bottomrule
\end{tabular}
}
\caption{Comparison of methods and baseline. Dual-Layer SNN and Ultralight SNN are marked as "D-L" and "U". LearnSTPLoss is marked as "w/L". (\textbf{The best}; \underline{The second best}.)}
\label{tab:main_results}
\end{table*}

\subsubsection{Temporal Consistency Error ($\mathrm{MSE}_T$).} The temporal error is computed by summing the squared differences of the voxel tensors across all spatiotemporal locations \cite{ES1}. Following our detailed definition, it is formally expressed as:
\begin{equation}
\mathrm{MSE}_T = \frac{1}{N_p} \sum_{i,j} \int_{T_0}^{T_1} \left( \text{Spike}_{i,j}^{h}(t) - \text{Spike}_{i,j}^{gt}(t) \right)^2 dt.
\end{equation}

\subsubsection{Spatial Consistency Error ($\mathrm{MSE}_S$).} To evaluate local event distribution over time, the event stream is divided into non-overlapping time blocks. For each block, a Peri-Stimulus Time Histogram (PSTH) is computed by summing voxel counts along the time dimension. $\mathrm{MSE}_S$ is defined as:

\begin{equation}
\mathrm{MSE}_S = \frac{1}{N_p} \sum_{k=1}^{N_b} \sum_{i,j} \left\| \text{PSTH}_{i,j}^{h}(k) - \text{PSTH}_{i,j}^{gt}(k) \right\|_2^2.
\end{equation}

\subsubsection{Root Mean Squared Error ($\mathrm{RMSE}_{\mathrm{ST}}$).}
The final RMSE score is defined as:
\begin{equation}
\mathrm{RMSE}_{\mathrm{ST}} = \sqrt{ \frac{1}{(T_1 - T_0) \cdot N_p} \left( \mathrm{MSE}_T + \mathrm{MSE}_S \right) }.
\end{equation}

\subsubsection{Polarity Accuracy (PA).} We also compute a new metric, the polarity accuracy (PA), defined as:
\begin{equation}
\text{PA} = \frac{ \left| \left\{ (x, y, t) \mid p^{\text{out}}_{x,y,t} = p^{\text{gt}}_{x,y,t} \right\} \cap \Omega \right| }{|\Omega|},
\end{equation}
where $\Omega$ denotes the set of shared spatiotemporal coordinates that are present in both the predicted and ground truth event streams. This metric reflects the model’s ability to preserve event polarity information, while, to the best of our knowledge, previous event super-resolution methods have not explicitly incorporated this metric.

\subsection{Comparative Results}
\subsubsection{Results of Super-resolved Event Stream.}
Table~\ref{tab:main_results} shows the comparison results between our methods and the baseline across five datasets. All four variants of our methods outperform the baseline on all event stream evaluation metrics, demonstrating the clear superiority of our dual-layer SNN architecture. Figure~\ref{fig:ASL} and Figure~\ref{fig:NMIST} show visualizations on the ASL-DVS and N-MNIST datasets, where our methods produce sharper details along dense edges and exhibit better suppression of edge-related noise. Figure \ref{fig:NFS} demonstrates the performance of our method on the EventNFS-real dataset, offering visual comparability with other event-to-frame super-resolution methods.

\begin{table}[t]
\centering
\resizebox{\linewidth}{!}{
\begin{tabular}{l|c|c|c}
\toprule
\textbf{Method} & \textbf{Params} & \textbf{FLOPs} & \textbf{Time} \\
\midrule
\textbf{SNN-based Event SR Methods} & & & \\
Dual-Layer SNN w/L (ours) & \underline{464 (0.464K)} & \textbf{0.53G} & \textbf{1.49ms} \\
Ultralight SNN w/L (ours) & \textbf{232 (0.232K)} & \underline{0.58G} & \underline{1.55\textsuperscript{\dag}ms} \\
Baseline \cite{ES1} & 1040 (1.04K) & 1.10G & 1.62ms \\
\midrule
\textbf{Traditional Image/Video SR Methods} & & & \\
SRFBN-esr \cite{SF5-21} & 2.1M (2100K) & 39.5G & 37.3ms \\
RLSP-esr \cite{SF5-8} & 1.2M (1200K) & 23.1G & -- \\
RSTT-esr \cite{SF5-13} & 3.8M (3800K) & 22.3G & 61.4ms \\
\midrule
\textbf{Event-to-Frame Event SR Methods} & & & \\
EventZoom \cite{EF3} & 11.5M (11500K) & 65.3G & 17.4ms \\
RecEvSR \cite{EF4} & 1.8M (1800K) & 2.80G & 13.2ms \\
BMCNET \cite{EF5} & 2.6M (2600K) & 35.35G & -- \\
RMFNET \cite{EF6} & 3.0M (3000K) & 8.73G & 7.0ms \\
\bottomrule
\end{tabular}
}
\caption{Comparison of model size, computational cost, and inference time on EventNFS dataset. \dag: Ultralight SNN uses concurrent dual forwards. Parts of the data are from \cite{EF5}.}
\label{tab:frame-based-evaluation}
\end{table}

\begin{figure*}[ht]
    \centering
    \includegraphics[width=\linewidth]{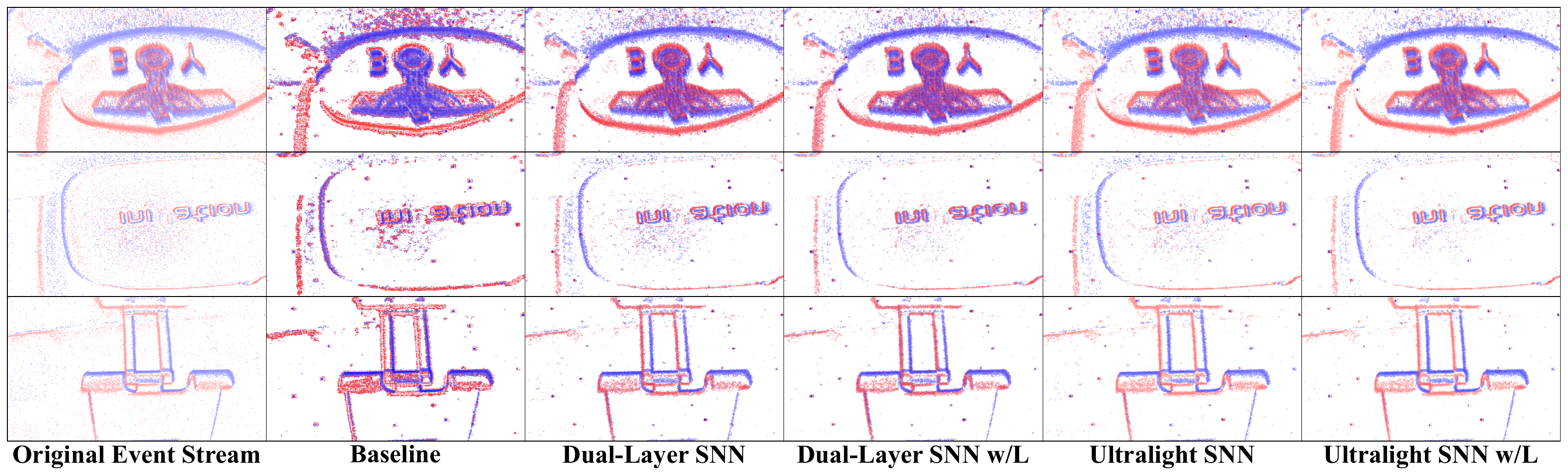}
    \caption{Visualization Results on Real-world Deployment.}
    \label{fig:DVX}
\end{figure*}

\subsubsection{Results of Model Lightweight.}
According to Table~\ref{tab:frame-based-evaluation}, our method achieves extreme lightweight efficiency. Compared to the baseline, the Ultralight SNN reduces the parameter count by 78\%. A trained Ultralight SNN on EventNFS-real occupies only 4.0KB of storage. Meanwhile, our inference speed surpasses all other methods, enabling the super-resolved event stream to appear on the monitor with virtually no perceptible delay from the scene.

\subsubsection{Results of Downstream Application.}
Referring Table~\ref{tab:image_recon}, we evaluate our method on the image reconstruction task using the Event Camera Dataset and E2VID \cite{E2VID}. The results show that our method improves SSIM by 7\% and reduces MSE by 4.48\% compared to the baseline. Figure~\ref{fig:image_recon} visualizes the reconstruction, where zoomed-up regions reveal that our method preserves clearer edge details. 

Additionally, we conduct classification tasks on N-MNIST, ASL-DVS, and CIFAR10-DVS datasets. Our method consistently improves the performance (See Appendix \cite{appendix}). These results also demonstrate that higher-resolution event data leads to better performance on these tasks, further validating the significance of super-resolving event data.

\begin{table}[t]
\centering
\resizebox{.9\linewidth}{!}{
\begin{tabular}{l|cc|cccc}
\toprule
\textbf{Metric} & \textbf{LR} & \textbf{HR-GT} & \textbf{(a)} & \textbf{Baseline} & \textbf{(b)} & \textbf{(c)} \\
\midrule
RMSE$_\text{ST}$ & -- & -- & 0.837 & 0.325 & \underline{0.313} & \textbf{0.312} \\
\midrule
SSIM$\uparrow$  & 0.460 & 0.624 & 0.497 & 0.530 & \underline{0.563} & \textbf{0.567} \\
MSE$\downarrow$      & 0.084 & 0.057 & 0.074 & 0.067 & \underline{0.066} & \textbf{0.064} \\
\bottomrule
\end{tabular}
}
\caption{Results of Image Reconstruction on Event Camera Dataset. (a) \cite{EF1} (b) Dual-Layer SNN w/L (c) Ultralight SNN w/L}
\label{tab:image_recon}
\end{table}

\subsubsection{Extended Validation.}
Since event-to-frame methods evaluate only the spatial quality of their frame-based HR outputs, our method is not directly comparable at the event stream level. In the appendix, we provide further analytical comparisons in terms of spatiotemporal quality. We also deploy a DVXplorer-Lite event camera in a mobile scenario to collect real-world data and evaluate our method, demonstrating its robustness in performing 2× super-resolution on raw event streams. The visualization results can be found in Figure \ref{fig:DVX}. (See Appendix \cite{appendix} for more details)

\subsection{Ablation Study}
According to Table~\ref{tab:main_results}, for the three spatio-temporal accuracy metrics, the Ultralight SNN outperforms the Dual-layer SNN on all datasets except N-MNIST. However, it achieves lower PA in three datasets. This suggests that the Dual-Forward strategy, by separating the learning and inference of positive and negative polarities, effectively mitigates spatio-temporal interference between polarities, resulting in better spatio-temporal super-resolution awareness, though at the cost of polarity estimation stability. In addition, when comparing SNNs trained with only spatial and temporal losses, the proposed LearnSTPLoss consistently improves nearly all evaluation metrics.

We also evaluated the Dual-Forward strategy and LearnSTPLoss individually on the baseline three-layer SNN using the N-MNIST dataset. As shown in Table~\ref{tab:ablation_study}, both components independently enhance the baseline performance, and their combination further improves the results.
\begin{table}[t]
\centering
\resizebox{\linewidth}{!}{%
\begin{tabular}{l|c|c|c|c}
\toprule
\textbf{Method} & RMSE$_\text{ST}$ & MSE$_\text{S}$ & MSE$_\text{T}$ & PA (\%) \\
\midrule
Baseline & 0.2720 & 0.1440 & 0.2301 & 96.36 \\
Baseline + DualForward & 0.2669 & \textbf{0.1389} & 0.2278 & 94.34 \\
Baseline + w/L & 0.2696 & 0.1415 & 0.2295 & \textbf{97.14} \\
Baseline + DualForward + w/L & \textbf{0.2661} & 0.1404 & \textbf{0.2260} & 96.12 \\
\bottomrule
\end{tabular}
}
\caption{Ablation study of Dual-Forward Strategy and LearnSTPLoss on N-MNIST with baseline method.}
\label{tab:ablation_study}
\end{table}

\section{Conclusion \& Future Work}
In conclusion, we present an ultra-lightweight, event-to-event stream-based super-resolution approach using SNNs. The proposed Dual-layer SNN structure and Dual-Forward Polarity-Split Event Encoding strategy enable separate processing of positive and negative events, reducing model size and computational cost while enhancing spatio-temporal consistency. Furthermore, the Learnable Spatio-temporal Polarity-aware Loss (LearnSTPLoss) adaptively balances spatial, temporal, and polarity fidelity, leading to improved reconstruction accuracy across diverse datasets. Experimental results demonstrate that our approach surpasses the baseline on five benchmark datasets and downstream tasks such as image reconstruction and object classification. Our method is the most lightweight and fastest in inference among all mainstream event-based super-resolution methods, making real-time, on-device deployment feasible.

In the future, we plan to deploy our model on real SNN hardware chips, such as Intel Loihi, to increase the temporal precision of spiking. We expect that this can further improve the results and speed, and will be more energy-saving.

\bibliography{main}

\clearpage
\appendix

\section*{Appendix}
\section{Real-world Deployment}

\begin{figure}[ht]
    \centering
    \includegraphics[width=\linewidth]{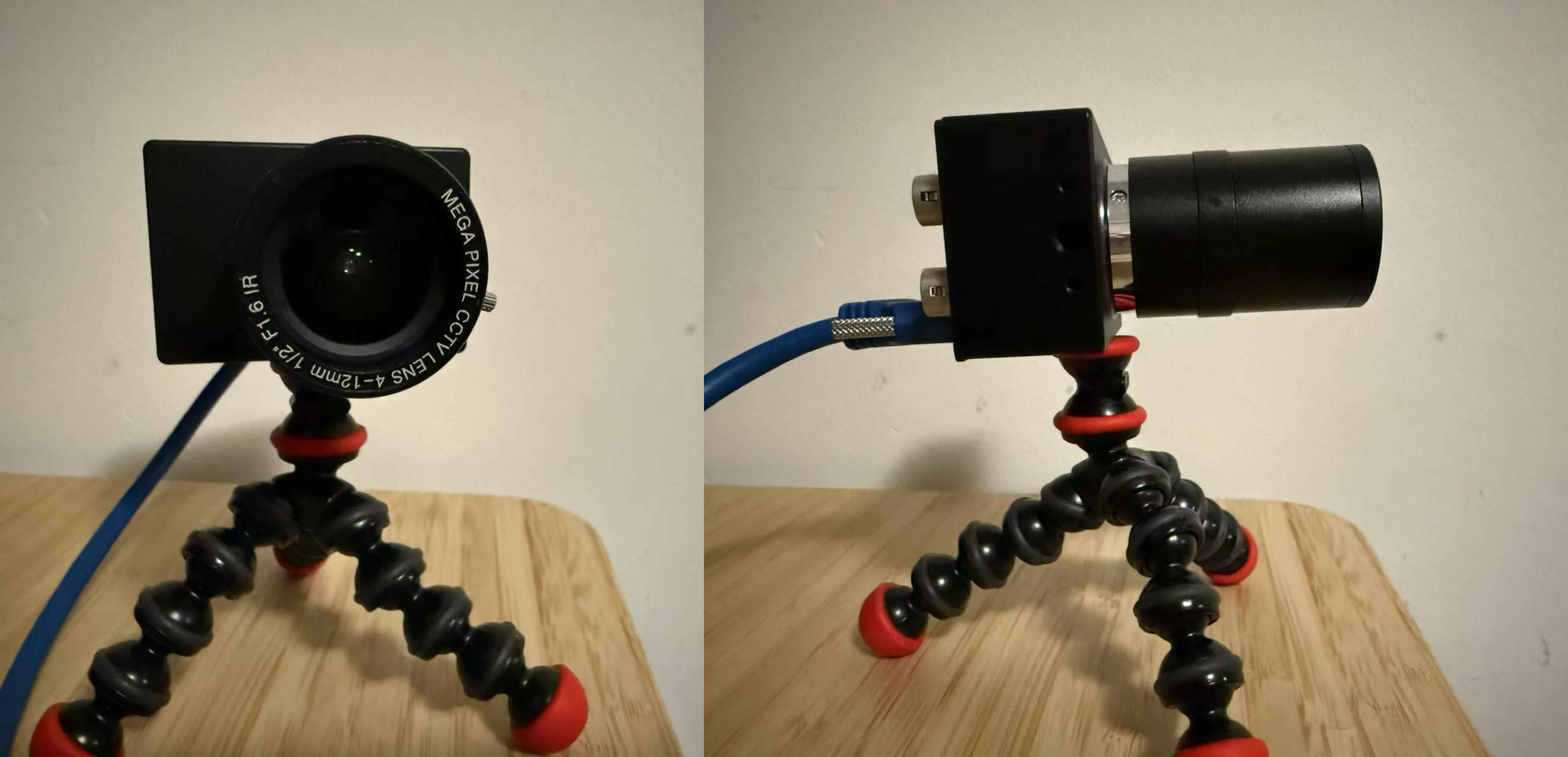}
    \caption{Our Real-world Deployment on Event Camera.}
    \label{fig:DVXplorer}
\end{figure}

\begin{figure*}[!ht]
    \centering
    \includegraphics[width=\linewidth]{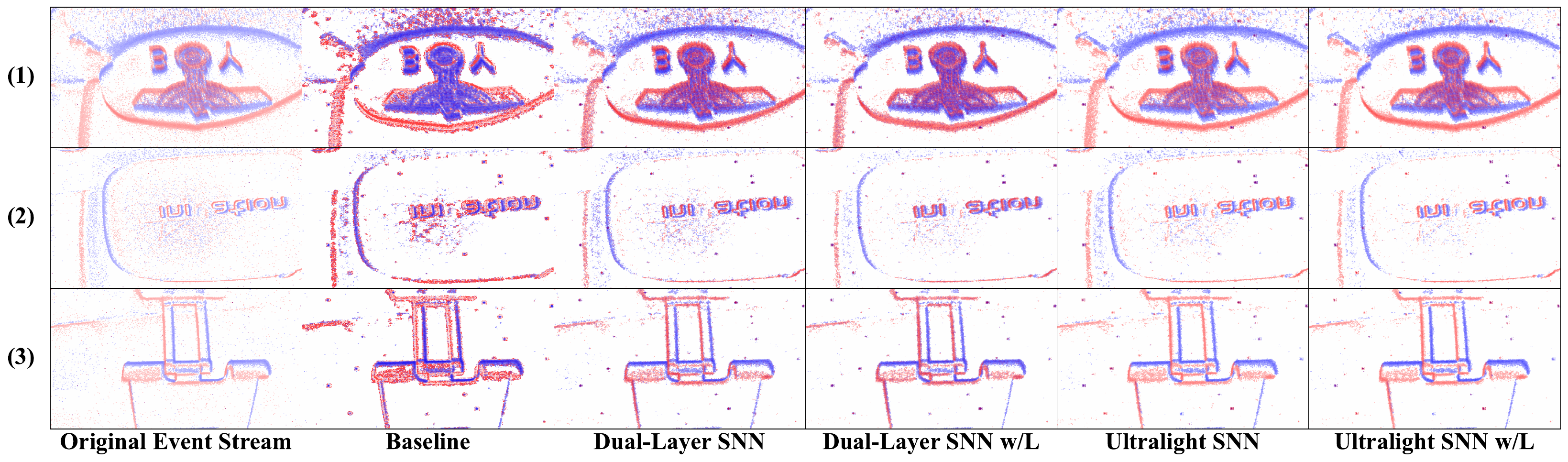}
    \caption{Visualization Results on Real-world Deployment.}
    \label{fig:DVX1}
\end{figure*}

To evaluate the robustness of our method in real-world scenarios, we deployed it in real time using a DVXplorer-Lite event camera shown in Figure \ref{fig:DVXplorer}, which captured data from some objects. We performed inference on the collected raw data using models trained on the EventNFS-real dataset for each method variant, and visualized the results, as shown in Figure~\ref{fig:DVX1}.

The visualization demonstrates the robustness of our method in real-world conditions, outperforming the baseline. It is worth noting that our super-resolution task can super-resolve the original resolution of $320 \times 240$ to $640 \times 480$, whereas the original event stream in EventNFS-real has a resolution of only $222 \times 124$. This highlights the robustness of our method in handling 2× super-resolution across different resolution domains.


\section{Supplementary Visualization Results}
\begin{figure*}[t]
    \centering
    \includegraphics[width=\linewidth]{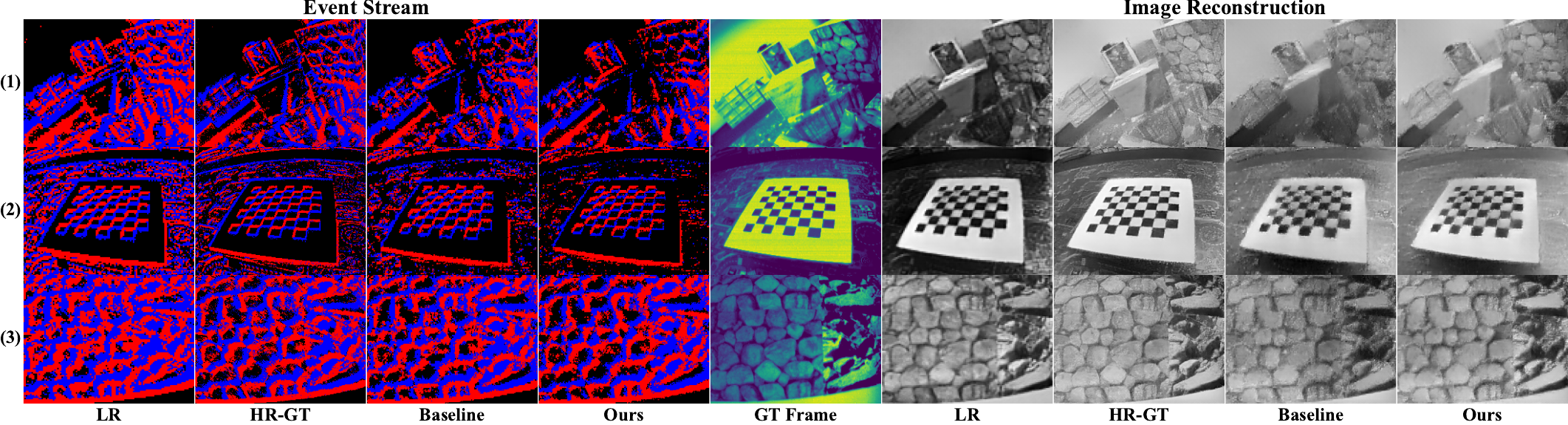}
    \caption{More Visualization Results on Event Camera Dataset.}
    \label{fig:ImageRecon}
\end{figure*}

\begin{figure*}[h]
    \centering
    \includegraphics[width=0.9\linewidth]{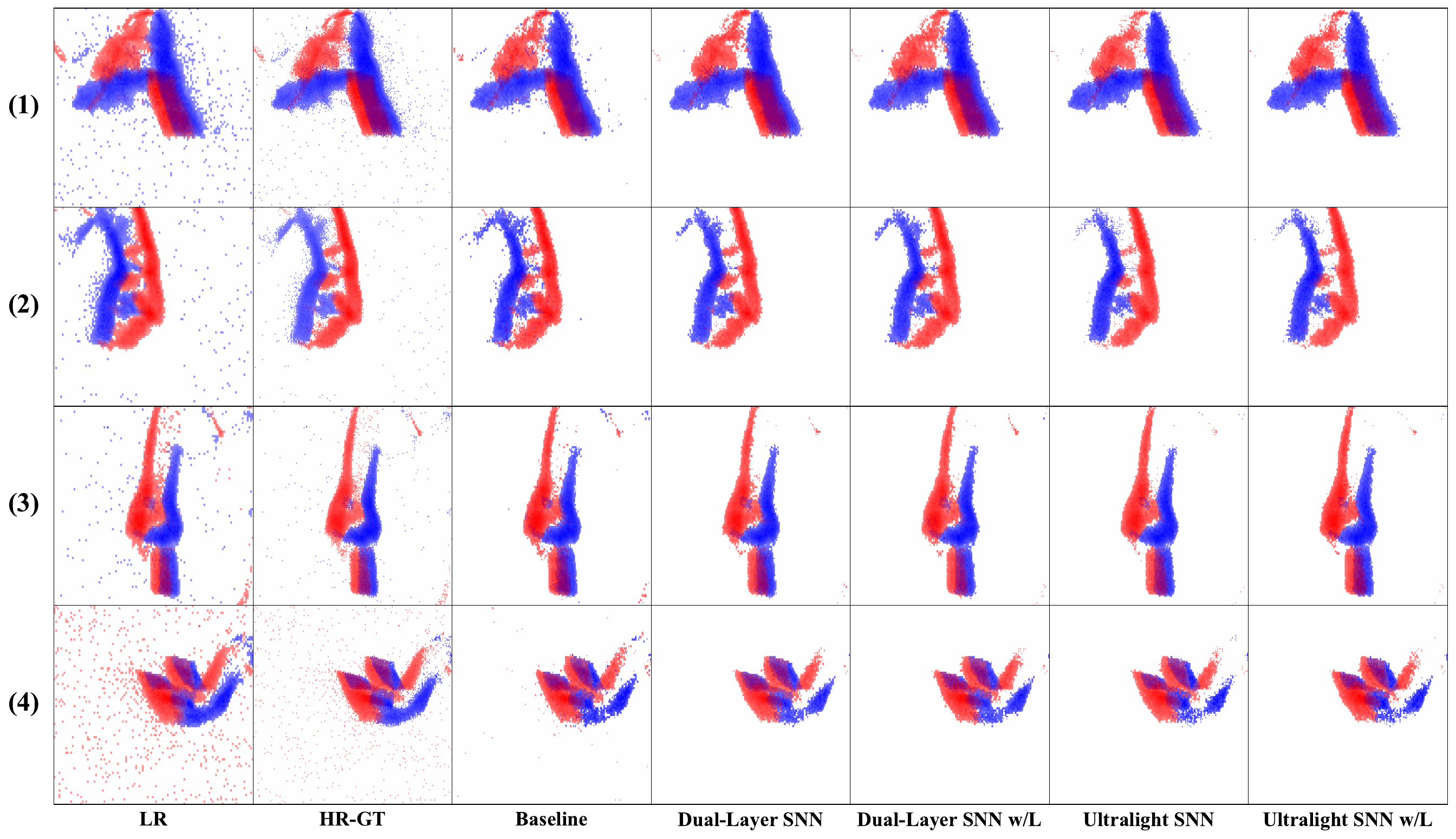}
    \caption{More Visualization Results on ASL-DVS}
    \label{fig:ASL1}
\end{figure*}

\begin{figure*}[h]
    \centering
    \includegraphics[width=0.9\linewidth]{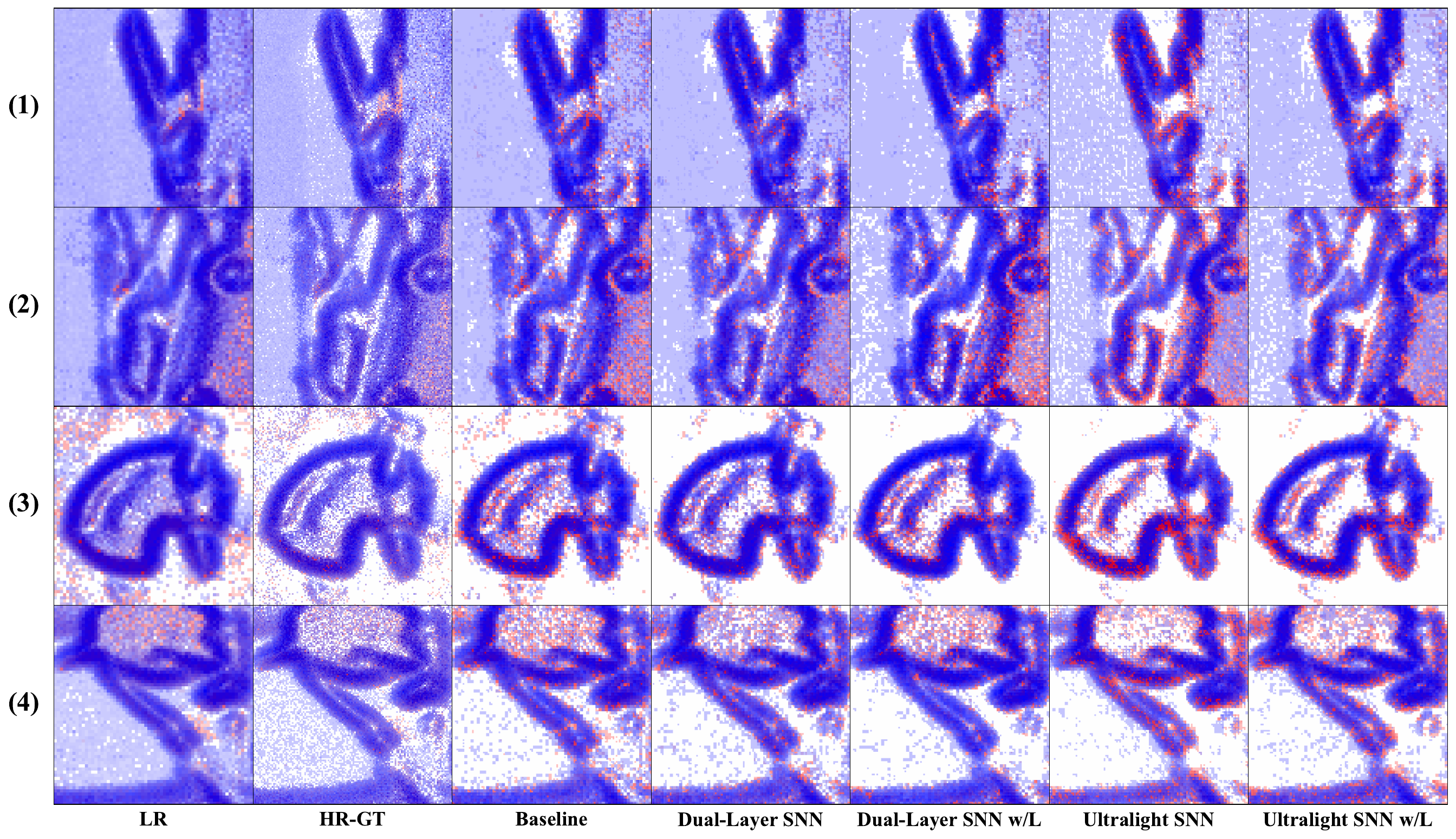}
    \caption{Visualization Results on CIFAR10-DVS}
    \label{fig:CIFAR}
\end{figure*}

We provide additional visual results to demonstrate the generalization capability of our method. Figure \ref{fig:ImageRecon} presents more event stream super-resolution and image reconstruction results on the Event Camera Dataset. Figure \ref{fig:ASL1} shows additional visualizations on the ASL-DVS dataset. Figure \ref{fig:CIFAR} provides visualization results on the CIFAR10-DVS dataset, which were not shown in the main paper.

\section{Supplementary Explanation for Event-to-Frame SR vs. Event-to-Event SR}
While both Event-to-Frame (E2F) and Event-to-Event (E2E) super-resolution methods aim to enhance the resolution of event streams, they differ fundamentally in data representation, network design, temporal modeling, and evaluation paradigms.

\textbf{Data Representation.}
E2F methods convert the original asynchronous event stream into synchronous frame-based forms, such as event count maps, event stacks, or voxel grids~\cite{EF1,EF3,EF6}. These representations discretize the temporal domain into fixed bins, allowing the use of standard CNNs designed for images or videos. Conversely, E2E methods directly process raw event tuples $(x,y,t,p)$ and preserve the asynchronous temporal structure throughout the network, usually requiring specialized architectures such as Spiking Neural Networks (SNNs)~\cite{ES1}.

\textbf{Temporal Modeling.}
E2F approaches sacrifice temporal resolution for spatial modeling flexibility. Temporal precision is either lost during accumulation or indirectly modeled via stacked inputs. E2E methods, on the other hand, retain event-level timestamps and employ explicit temporal constraints (e.g., spike train losses or inter-frame consistency) to learn both fine-grained timing and spatial detail.

\textbf{Network Architecture and Output.}
Due to their reliance on dense grid-based inputs and outputs, E2F networks often use complex architectures with large parameter counts (e.g., 3D U-Nets, RNNs, Transformers), making them computationally expensive and harder to deploy on neuromorphic devices. E2E methods typically adopt lightweight spiking architectures tailored for sparse asynchronous data, which are more suitable for real-time deployment on event cameras.

\textbf{Evaluation.}
Crucially, the evaluation metrics for E2F methods are designed for dense frames and cannot reflect temporal fidelity or spike timing accuracy. Since E2E methods produce the event stream, comparing their output with dense E2F frames using frame-based metrics is inherently unfair. Thus, the two types of methods operate under different assumptions and are designed for different downstream tasks. Any direct numerical comparison would be misleading unless both outputs are projected to a common representation with equal temporal resolution, which itself introduces biases.

In summary, while E2F and E2E SR share the same goal, they are structurally and functionally different. E2F methods prioritize spatial resolution with frame-based reasoning, whereas E2E methods aim to reconstruct temporally precise asynchronous data.

\section{Supplementary Explanation for Methods}
\subsection{Spiking Neural Networks}
Spiking Neural Networks (SNNs) are considered the third generation of neural networks, offering a biologically plausible framework for processing temporal data in a highly efficient and event-driven manner~\cite{maass1997networks}. Unlike conventional artificial neural networks (ANNs) that rely on continuous-valued activations, SNNs communicate through discrete spikes, enabling sparse and asynchronous computations that closely mimic the behavior of real biological neurons. Each neuron in an SNN integrates incoming spikes over time through a membrane potential and generates an output spike only when this potential exceeds a certain threshold~\cite{gerstner2002spiking}.

This temporal coding scheme allows SNNs to naturally model time-dependent dynamics, making them especially suitable for neuromorphic sensors such as event cameras, which output asynchronous streams of spatiotemporal events. Additionally, SNNs have demonstrated promising energy efficiency when deployed on neuromorphic hardware like Intel's Loihi~\cite{davies2018loihi} or the SpiNNaker platform~\cite{furber2014spinnaker}, due to their sparse activity and low-power operation. However, training SNNs remains challenging because the non-differentiable nature of spike generation complicates the use of standard backpropagation. To address this, surrogate gradient methods~\cite{neftci2019surrogate} and ANN-to-SNN conversion techniques~\cite{rueckauer2017conversion} have been widely studied, enabling SNNs to learn complex tasks with performance approaching that of traditional deep learning models.

\subsection{SRM-based Spiking Neuron Encoding (Extended Explanation)}

In this section, we provide a detailed description of the SRM-based spiking neuron dynamics used in our SNN architecture for event stream super-resolution \cite{SRM}. Our model leverages temporal encoding and biological realism to effectively process asynchronous spike events.

\paragraph{Post-Synaptic Potential (PSP).}
The \textit{post-synaptic potential} (PSP) represents the transient response of a neuron's membrane potential to incoming spikes. For each input spike train $s_\text{in}^{(i)}(t)$, the influence is computed via a convolution with the spike response kernel $\epsilon(t)$:
\[
\text{PSP}^{(i)}(t) = \epsilon(t) * s_\text{in}^{(i)}(t)
\]
\[
\epsilon(t) = \frac{t}{\tau_s} \exp\left(1 - \frac{t}{\tau_s}\right) \cdot \Theta(t)
\]
where $\tau_s$ is the synaptic time constant and $\Theta(t)$ is the Heaviside step function to ensure causality.

\paragraph{Membrane Potential and Spiking.}
The membrane potential $u(t)$ integrates PSPs from multiple inputs, weighted by synaptic weights $w^{(i)}$, along with a refractory suppression term:
\[
u(t) = \sum_i w^{(i)} \cdot \text{PSP}^{(i)}(t) + \gamma(t) * s_\text{out}(t)
\]
The refractory kernel $\gamma(t)$, modeling post-spike inhibition, is defined as:
\[
\gamma(t) = -\lambda \cdot \exp\left(-\frac{t}{\tau_r}\right) \cdot \Theta(t)
\]
where $\lambda$ is a suppression coefficient and $\tau_r$ is the refractory time constant.

When $u(t)$ exceeds the threshold $V_\text{th}$, the neuron emits a spike:
\[
s_\text{out}(t) = \sum_{t_k \in \{ t \mid u(t) = V_\text{th} \}} \delta(t - t_k)
\]

This membrane potential typically starts at a constant resting potential $V_{rest}$, often set to 0 in simplified models, representing the neuron’s state in the absence of input.
\paragraph{Feed-forward Dynamics.}
For an $L$-layer feed-forward SNN, the membrane potential and output spike train of the $(l+1)$-th layer are:
\begin{align*}
u^{(l+1)}(t) &= \mathbf{W}^{(l)} \cdot \left[ \epsilon(t) * s^{(l)}(t) \right] + \gamma(t) * s^{(l+1)}(t) \\
s^{(l+1)}(t) &= \sum_{t_k \in \{t \mid u^{(l+1)}(t) = V_\text{th} \}} \delta(t - t_k)
\end{align*}

\paragraph{Biological Motivation.}
This SRM-based neuron formulation enables:
\begin{itemize}
    \item \textbf{Temporal integration} of input signals through PSP.
    \item \textbf{Refractory suppression} without membrane reset.
    \item \textbf{Sparsity and energy efficiency}, benefiting from asynchronous spike-based computation.
\end{itemize}

These features make SRM-based SNNs naturally compatible with event-based data and well-suited for tasks requiring temporal precision, such as event stream super-resolution.

\subsection{Concurrent Dual-Forward SNN}
In the main paper, we propose a neuromorphic forward propagation strategy named \textbf{Dual-Forward Polarity-split Event Encoding}, which explicitly separates positive and negative polarity events into two independent data streams and performs two forward passes. This design enhances spatio-temporal fidelity while maintaining modularity and compatibility with lightweight architectures.

To better adapt this strategy to real deployment scenarios, especially on resource-constrained neuromorphic or edge devices, we introduce two controllable execution modes for the Ultralight SNN:

\begin{itemize}
    \item \textbf{Sequential Dual Inference}

 In this mode, the network performs two forward passes for positive and negative polarities \textbf{sequentially}. This approach is particularly suitable for devices with low power consumption or limited memory, as the memory footprint and computational load at any moment are halved compared to the concurrent mode. Although this results in slightly increased inference latency, it is practical for embedded or mobile systems without parallel computing capabilities.
    \item \textbf{Concurrent Dual Inference}

 In this mode, the network processes positive and negative polarity streams \textbf{concurrently} using separate computational threads or processing units. This enables full exploitation of parallelism on CPUs, resulting in minimal inference latency. Since both streams share the same network weights and structure, learning consistency is ensured while significantly improving execution efficiency. This mode is highly suitable for high-performance scenarios with strict speed requirements.
\end{itemize}
\begin{table}[h]
\centering
\label{tab:inference_time}
\begin{tabular}{l|c}
\hline
\textbf{Method} & \textbf{Time (ms)} \\
\hline
Dual-Layer SNN & 1.49 \\
Ultralight SNN (\textbf{Sequential}) & 3.71 \\
Ultralight SNN (Concurrent) & 1.55 \\
Baseline & 1.62 \\
\hline
\end{tabular}
\caption{Average inference time per EventNFS-real sample on CPU.}
\label{inference_time}
\end{table}
Both modes maintain \textbf{identical output structures and supervision signals}, thus allowing flexible deployment depending on hardware availability. Table \ref{inference_time} shows the average inference time per EventNFS-real data sequence on the CPU.

\subsection{PSP-based Residual Upsampling Bypass}
To enhance the reconstruction fidelity while preserving the compactness of the Dual-layer SNN architecture, we use a residual upsampling bypass branch based on post-synaptic potential (PSP). This bypass provides a non-spiking shortcut path from the low-resolution input to the high-resolution output, complementing the main spiking computation pipeline.

Specifically, after applying the first spiking convolution layer and PSP transformation, the resulting membrane potential tensor $\texttt{psp}$ of shape $[B, C, H, W, T]$ is reshaped and upsampled in the spatial domain using bilinear interpolation. This produces a tensor $\texttt{psp\_bypass}$ with the same spatial resolution as the desired output but without spiking discretization. This upsampled potential retains the continuous temporal and spatial dynamics of the original low-resolution events. It is then added to the upsampled spike response from the main pathway before the final spike activation:
\begin{align*}
s_\text{out} = \texttt{spike}\left( \texttt{upconv}\left( \texttt{psp}(s_1) \right) + \texttt{psp\_bypass} \right),
\end{align*}
where $s_1$ is the intermediate spike output from the first spiking layer.

Compared to the main spiking pathway, which involves quantized spike-based transmission and nonlinear neuron thresholds, the PSP-based bypass maintains a smoother, continuous approximation of the input signal. This helps preserve fine-grained edge and texture information that may be lost during spike quantization, especially under sparse input conditions. Such a design aligns with similar principles found in residual learning~\cite{He2016ResNet}, where shortcut connections assist in gradient propagation and retain low-level details.

Importantly, this bypass does not require additional trainable parameters or spiking computation, thus preserving the model's lightweight property while improving the reconstruction quality. The combination of spike-driven dynamics and PSP-based residual flow enables the model to strike a balance between biological plausibility and practical performance in event stream super-resolution.

\section{Supplementary Explanation for Experiments}

\subsection{Calculation of $MSE_S$ and $MSE_T$}

To complement the evaluation section, we provide further details on the implementation and interpretation of the two components of the event stream RMSE metric: the \textbf{temporal error (MSE$_T$)} and \textbf{spatial error (MSE$_S$)}.

\paragraph{Temporal Consistency Error ($\mathrm{MSE}_T$).}
This term evaluates the voxel-wise difference over the full spatiotemporal domain between the predicted and ground-truth event streams. As shown in Eq.~(7), both streams are discretized into 4D voxel grids of shape $(2, H, W, T)$, where the two channels represent positive and negative polarities. $\mathrm{MSE}_T$ is computed as the squared difference between corresponding voxels over time, capturing pixel-level spike misalignment:
\[
\mathrm{MSE}_T = \frac{1}{N_p} \sum_{i,j} \int_{T_0}^{T_1} \left( \text{Spike}_{i,j}^{h}(t) - \text{Spike}_{i,j}^{gt}(t) \right)^2 dt.
\]
In our implementation, this integral is approximated by voxel-wise summation over the discretized tensor using PyTorch.

\paragraph{Spatial Consistency Error ($\mathrm{MSE}_S$).}
To evaluate local event distribution over time, the event stream is divided into non-overlapping time blocks (e.g., 50 ms). For each block, a Peri-Stimulus Time Histogram (PSTH) is computed by summing voxel counts along the time dimension. $\mathrm{MSE}_S$ is defined as:
\[
\mathrm{MSE}_S = \frac{1}{N_p} \sum_{k=1}^{N_b} \sum_{i,j} \left\| \text{PSTH}_{i,j}^{h}(k) - \text{PSTH}_{i,j}^{gt}(k) \right\|_2^2.
\]
This captures the spatial mismatch of event accumulation per pixel in each time bin.

\paragraph{Normalization.}
The final RMSE is normalized by both the number of active pixels $N_p$ and the total temporal span $(T_1 - T_0)$:
\[
\mathrm{RMSE}_{\mathrm{ST}} = \sqrt{ \frac{1}{(T_1 - T_0) \cdot N_p} \left( \mathrm{MSE}_T + \mathrm{MSE}_S \right) }.
\]

\paragraph{Implementation Notes.}
Our implementation follows the above formulation. Voxel tensors are constructed using indexed PyTorch operations. $\mathrm{MSE}_T$ and $\mathrm{MSE}_S$ are computed via tensor summations. Additionally, a \textbf{polarity accuracy (PA)} metric is calculated by intersecting spatiotemporal coordinates of predicted and ground-truth events, then comparing their polarities:
\[
\text{PA} = \frac{ \left| \left\{ (x, y, t) \mid p^{\text{out}}_{x,y,t} = p^{\text{gt}}_{x,y,t} \right\} \cap \Omega \right| }{|\Omega|},
\]
where $\Omega$ is the set of shared coordinates between prediction and ground truth. This metric reflects the model's ability to preserve correct polarity.

\subsection{Calculation of GFLOPS for SNN}

To estimate the computational cost of our SNN architecture, we report the theoretical number of floating-point operations (FLOPs). As our network is implemented using the SlayerSNN library, which simulates spiking dynamics via time-unfolded convolutional operations, the FLOPs count accumulates across both spatial and temporal dimensions.

Let the input be a 5D spike tensor of shape $(B, C, H, W, T)$, where $B$ is the batch size, $C$ is the input channel count (2 for positive/negative polarity), and $T$ is the number of discrete timesteps. For a standard convolutional or transposed convolutional layer in SlayerSNN, the total FLOPs can be approximated as:
\begin{align*}
\mathrm{FLOPs}_\text{layer} = 2 \cdot K_H \cdot K_W \cdot C_\text{in} \cdot C_\text{out} \cdot H_\text{out} \cdot W_\text{out} \cdot T,
\end{align*}
where $(K_H, K_W)$ is the kernel size, $C_\text{in}$ and $C_\text{out}$ are the input/output channels, $(H_\text{out}, W_\text{out})$ is the output spatial size, and the factor of 2 accounts for multiply-accumulate (MAC) operations.

For our \texttt{Dual-layer SNN}, which contains:
\begin{itemize}
  \item A convolutional layer: $\texttt{conv1}$ with kernel size $5\times5$, input channel $2$, output channel $8$, spatial size $(H, W)$, and temporal length $T$.
  \item A transposed convolution: $\texttt{upconv1}$ with kernel size $2\times2$, input channel $8$, output channel $2$, spatial size $(2H, 2W)$, and the same $T$.
\end{itemize}

The total FLOPs can be estimated as:
\begin{align*}
\mathrm{FLOPs}_\text{conv1} &= 2 \cdot 5 \cdot 5 \cdot 2 \cdot 8 \cdot H \cdot W \cdot T, \\
\mathrm{FLOPs}_\text{upconv1} &= 2 \cdot 2 \cdot 2 \cdot 8 \cdot 2 \cdot (2H) \cdot (2W) \cdot T.
\end{align*}

The total GFLOPS is computed by summing the above and dividing by $10^9$:
\begin{align*}
\mathrm{GFLOPs} = \frac{\mathrm{FLOPs}_\text{conv1} + \mathrm{FLOPs}_\text{upconv1}}{10^9}.
\end{align*}

Therefore, the GFLOPs calculation result for \texttt{Dual-layer SNN} is around 0.531G.

This calculation does not include PSP, spike generation, or interpolation operations, which are typically lightweight but time-unfolded. Due to the use of non-standard temporal modules (e.g., snn.layer, psp, spike) and time-dependent membrane updates defined in the slayerSNN library, our model is incompatible with popular FLOPs estimation tools like ptflops and torchinfo.

\subsection{Downstream Application: Object Classification}
As described in the main paper, in addition to image reconstruction, we also evaluated the performance of the super-resolved event data on an object classification task. We employed the Event Spike Tensor-based event classification method proposed by Gehrig et al. \cite{Classification} for this evaluation. The experimental results are presented in Table~\ref{Classification}. Since the baseline method has already achieved the highest classification accuracy, equivalent to that of original high-resolution events, on the N-MNIST and ASL-DVS datasets, our method maintains the same performance. However, on the CIFAR10-DVS dataset, our method shows a slight improvement.

\begin{table}[t]
\centering
\resizebox{\linewidth}{!}{
\begin{tabular}{l|ccc}
\toprule
\textbf{Method} & \textbf{N-MNIST} & \textbf{ASL-DVS} & \textbf{CIFAR10-DVS} \\
\midrule
LR events         & 99.0\%  & 99.7\%  & 50.8\% \\
Li \textit{et al.}~\cite{EF1} & 97.8\%  & 98.0\%  & 59.6\% \\
Baseline          & 99.1\%  & 99.9\%  & 76.8\% \\
Dual-layer SNN    & 99.1\%  & 99.9\%      & 77.1\%     \\
Ultralight SNN    & 99.1\%  & 99.9\%      & 77.3\%    \\
\midrule
GT events (ref.)  & 99.1\%  & 99.9\%  & 78.7\% \\
\bottomrule
\end{tabular}
}
\caption{Classification accuracy (\%) on three event datasets.}
\label{Classification}
\end{table}

\subsection{Distribution of Neuromorphic Features before and after SR}
\begin{figure*}[ht]
    \centering
    \begin{subfigure}{\textwidth}
        \centering
        \includegraphics[width=\textwidth]{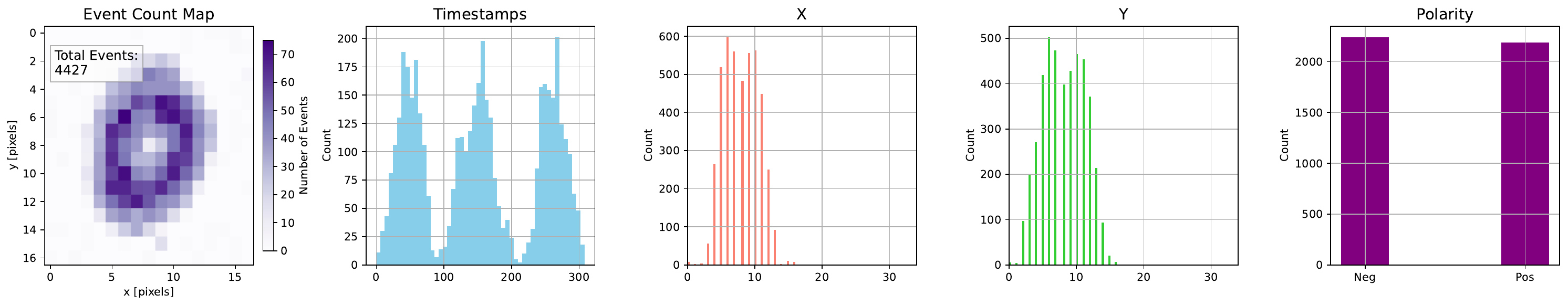}
        \caption{LR Event Stream.}
        \label{fig:sub1}
    \end{subfigure}
    \vspace{0.5em}
    
    \begin{subfigure}{\textwidth}
        \centering
        \includegraphics[width=\textwidth]{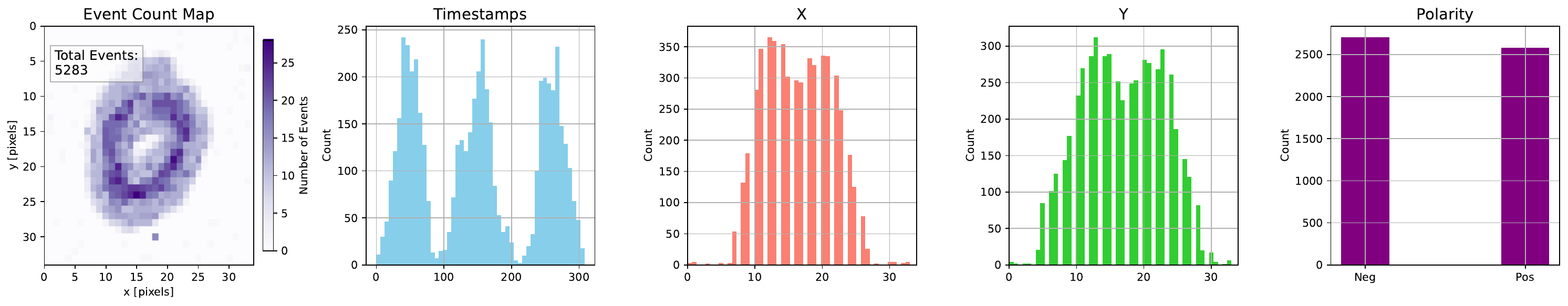}
        \caption{HR Event Stream.}
        \label{fig:sub2}
    \end{subfigure}
    \vspace{0.5em}

    \begin{subfigure}{\textwidth}
        \centering
        \includegraphics[width=\textwidth]{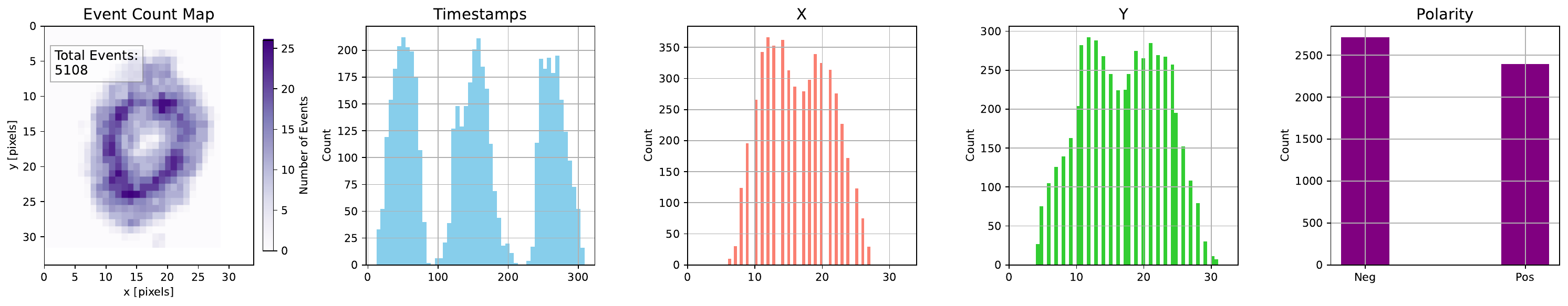}
        \caption{Event Stream from Dual-layer SNN.}
        \label{fig:sub3}
    \end{subfigure}
    \vspace{0.5em}

    \begin{subfigure}{\textwidth}
        \centering
        \includegraphics[width=\textwidth]{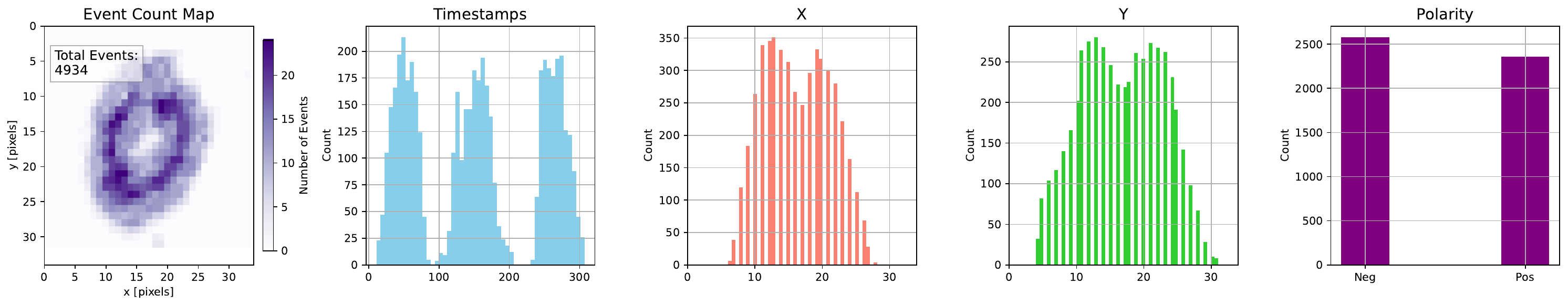}
        \caption{Event Stream from Ultralight SNN.}
        \label{fig:sub4}
    \end{subfigure}

    \caption{Spatiotemporal and polarity distributions of event stream before and after super-resolution.}
    \label{fig:distribution}
\end{figure*}

To demonstrate the nature of our event super-resolution method, we visualize the spatiotemporal and polarity distributions of a randomly selected digit-0 sample from the N-MNIST dataset before and after super-resolution, as shown in Figure \ref{fig:distribution}.
Most notably, the spatial resolution is doubled along both the X and Y dimensions, which is clearly reflected in the Event Count Map.

Secondly, due to spatial downsampling, four neighbouring pixels in the HR event stream are merged into one in the LR stream, resulting in fewer events in the LR version. Our method addresses this by not only upsampling the spatial resolution but also reasonably increasing the number of events to reflect the original density, rather than simply applying full or random interpolation. Our method tends to generate slightly fewer events than the HR ground truth, while maintaining accurate spatial reconstruction, which is a behavior we consistently observe across various samples.
It is also worth noting that, based on extensive visual inspection, the missing events in our super-resolved stream often correspond to noisy or isolated outliers in the HR stream, suggesting our method may also exhibit a denoising effect.

Finally, in the temporal domain, our super-resolved event stream preserves asynchronous and continuous timestamps. The distribution of events over time remains largely unchanged before and after super-resolution, indicating that our model effectively maintains temporal fidelity while enhancing spatial resolution.

\end{document}